\pdfoutput=1

\documentclass[11pt]{article}

\usepackage[final]{acl}

\usepackage{times}
\usepackage{latexsym}
\usepackage{multicol}
\usepackage{tabularx}
\usepackage{geometry}
\usepackage{booktabs}
\usepackage{caption}
\usepackage{afterpage}
\usepackage{float}
\usepackage{stfloats}
\usepackage{amsmath}
\usepackage{amssymb}
\usepackage{enumitem}

\usepackage[T1]{fontenc}

\usepackage[utf8]{inputenc}

\usepackage{microtype}

\usepackage{inconsolata}

\usepackage{graphicx}
\usepackage{amsmath}
\title{Benchmarking Large Language Models on Multiple Tasks in Bioinformatics NLP with Prompting}

\author{
\bf Jiyue Jiang$^{\heartsuit}$$^{\ast}$,
Pengan Chen$^{\spadesuit}$\thanks{Equal Contribution},
Jiuming Wang$^{\heartsuit}$, 
Dongchen He$^{\heartsuit}$, 
Ziqin Wei$^{\heartsuit}$, 
Liang Hong$^{\heartsuit}$, \\
\bf Licheng Zong$^{\heartsuit}$, 
Sheng Wang$^{\spadesuit}$, 
Qinze Yu$^{\heartsuit}$, 
Zixian Ma$^{\heartsuit}$, 
Yanyu Chen$^{\heartsuit}$, 
Yimin Fan$^{\heartsuit}$, \\
\bf Xiangyu Shi$^{\heartsuit}$,
Jiawei Sun$^{\diamondsuit}$, 
Chuan Wu$^{\spadesuit}$, 
Yu Li$^{\heartsuit}$ \\
$^{\heartsuit}$ The Chinese University of Hong Kong, $^{\spadesuit}$ The University of Hong Kong, $^{\diamondsuit}$ Shanghai AI Lab \\
{\tt
\{jiangjy, jmwang, dche, 1155173761, liang.hong, lczong, 1155183728, 1155191449, } \\
{\tt
fanyimin\}@link.cuhk.edu.hk, 
\{cpa2001, u3009618\}@connect.hku.hk, \{charliegood2019, } \\
{\tt
sxysxygm\}@gmail.com, 
sunjiawei1@pjlab.org.cn,
cwu@cs.hku.hk,
liyu@cse.cuhk.edu.hk
}
}

\begin{document}
\maketitle
\begin{abstract}

Large language models (LLMs) have become important tools in solving biological problems, offering improvements in accuracy and adaptability over conventional methods. Several benchmarks have been proposed to evaluate the performance of these LLMs. However, current benchmarks can hardly evaluate the performance of these models across diverse tasks effectively. In this paper, we introduce a comprehensive prompting-based benchmarking framework, termed Bio-benchmark, which includes 30 key bioinformatics tasks covering areas such as proteins, RNA, drugs, electronic health records, and traditional Chinese medicine. Using this benchmark, we evaluate six mainstream LLMs, including GPT-4o and Llama-3.1-70b, etc., using 0-shot and few-shot Chain-of-Thought (CoT) settings without fine-tuning to reveal their intrinsic capabilities. To improve the efficiency of our evaluations, we demonstrate BioFinder, a new tool for extracting answers from LLM responses, which increases extraction accuracy by round 30\% compared to existing methods. Our benchmark results show the biological tasks suitable for current LLMs and identify specific areas requiring enhancement. Furthermore, we propose targeted prompt engineering strategies for optimizing LLM performance in these contexts. Based on these findings, we provide recommendations for the development of more robust LLMs tailored for various biological applications. This work offers a comprehensive evaluation framework and robust tools to support the application of LLMs in bioinformatics.

\end{abstract}

\section{Introduction}

Computational methods have become essential for solving many biological problems, such as protein folding~\cite{jumper2021highly}, function annotation~\cite{singh2024llms}, and designing new biomolecules~\cite{lisanza2024multistate}. With the rise of language models in natural language processing (NLP)~\cite{brown2020language, jiang2023cognitive}, especially LLMs~\cite{touvron2023llama, bubeck2023sparks, guo2025deepseek}, these powerful tools can now be applied to biology as well, especially models with over 1 billion parameters, such as ProLLaMA~\cite{lv2024prollama}, Evo~\cite{nguyen2024sequence, evo2}, and Med-PaLM 2~\cite{singhal2023towards}. 
In addition to handling biomedical text data such as electronic health record (EHR) and traditional Chinese medicine (TCM) question-answering (QA) systems, LLMs are also effective in analyzing biological targets like proteins and nucleic acids due to their similarity to natural language sequences. Utilizing LLMs in biological tasks has resulted in significant advancements over traditional methods, including improved accuracy, better generalization, and enhanced learning from fewer samples~\cite{zhang2024scientific}.

\begin{figure*}[t]
    \centering
    \includegraphics[width=13cm]{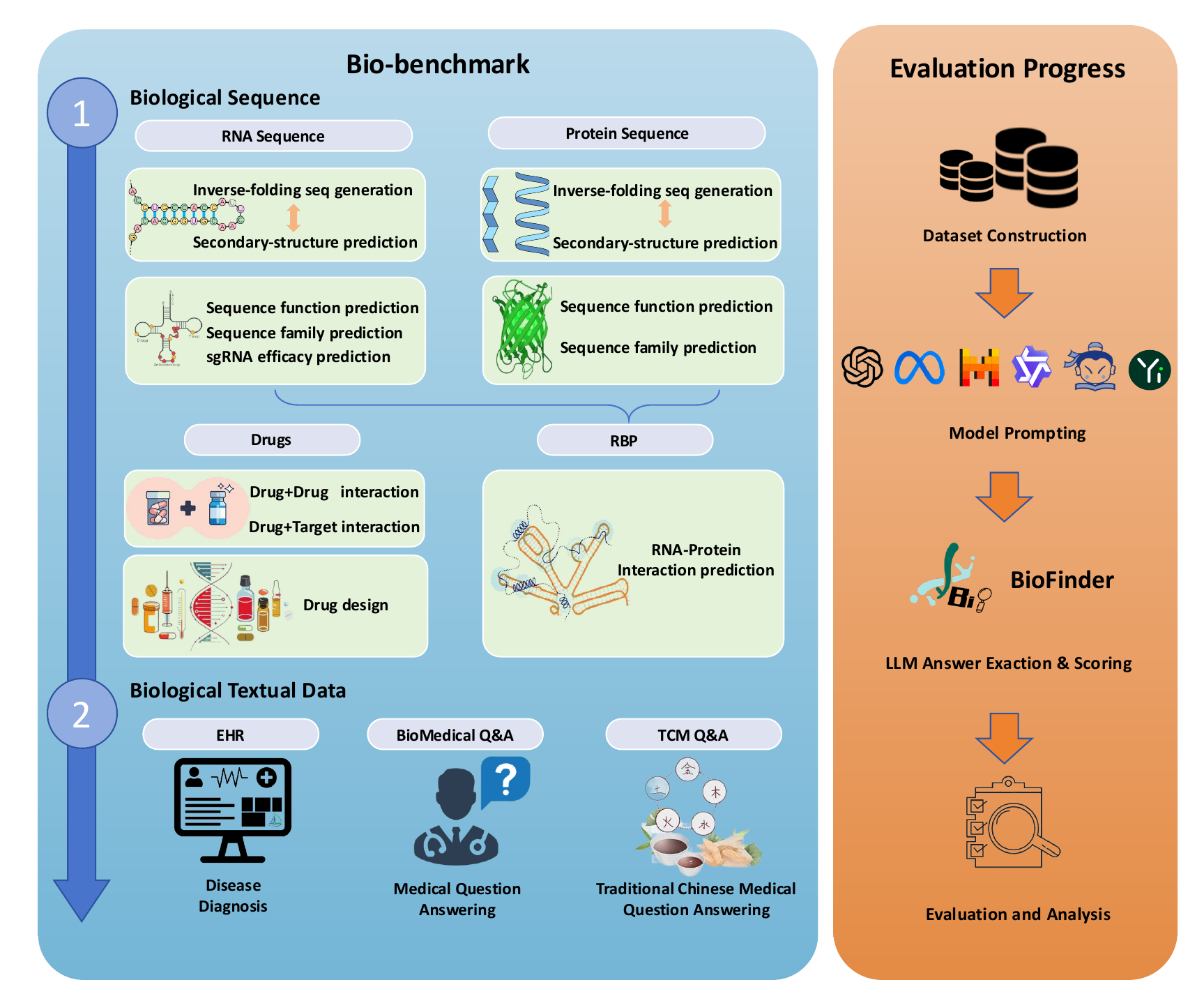}
    \caption{Overview of the paper. Bio-benchmark is divided into sequence (Protein, RNA, RBP, Drug) and text (EHR, Medical, TCM) data (left). The process is to use six representative LLMs to generate answers for a total of 30 subtasks in seven domains in biological information through 0-shot and few-shot after the benchmark is built. After extracting the answers generated by LLMs using our proposed BioFinder, we evaluate and analyze the extracted answers of different LLMs (right). \textbf{The experiment results table is in Appendix Table~\ref{complete}, Figure~\ref{bb1},~\ref{bb2}, and~\ref{res}.}}
    \label{fig:modeloverview}
\end{figure*}

A key strategy for leveraging LLMs in biology is prompting, which involves designing specific input formats to guide the model’s output~\cite{liu2023pre}. Prompting offers several benefits: it requires less computational power and data compared to fine-tuning, enables rapid adaptation to various tasks, and fully utilizes the model’s existing knowledge. Additionally, prompting facilitates the resolution of complex biological questions by framing them in a way that LLMs can effectively interpret and answer.

Many specialized LLMs have been developed for different biological tasks~\cite{wang2024multi,lv2024prollama,bolton2024biomedlm}, but there is currently no thorough investigation of which biological task is better suited for which LLM architecture.
Benchmarking is a rigorous method for assessing the compatibility and performance of an LLM architecture across a variety of tasks~\cite{mcintosh2024inadequacies, jiang2024well, ali2024using}.
However, current benchmark methods for LLMs on biological tasks face several challenges. 
First, most benchmarks designed for small-scale models are inadequate for evaluating LLMs due to their expanded hypothesis space. 
Second, existing validation datasets often contain overlapping data, highlighting the necessity for high-quality, clean benchmark sets. 
Third, despite the growing interest in developing LLMs for diverse tasks, there are limited benchmarks tailored to evaluate these comprehensive models.

Inspired by these needs, this paper introduces a comprehensive prompting-based benchmarking scheme for evaluating the performance of LLMs on various biological tasks. 
We present evaluation methods for LLMs using either biological sequence data or biomedicine-related text inputs, and apply these benchmarks to currently available LLMs, with an emphasis on general LLMs such as GPT-4o.
Our assessment specifically focuses on benchmarking vanilla LLMs without fine-tuning under zero-shot or few-shot CoT~\cite{wei2022chain} conditions to test the intrinsic abilities of LLMs, which can also better inform their performance after fine-tuning. 
Moreover, as LLMs often embed the key answer within the generated text output and there are currently no effective methods for extracting biological sequences and texts~\cite{gu2022robustness}, we propose a novel answer extraction method, BioFinder, to retrieve key answers from responses and enhance the reliability of our LLM benchmark.
Based on our comprehensive benchmark results, we analyze and summarize the biological tasks that are suitable for current LLMs to solve and provide architectural recommendations to guide the development of future LLMs for diverse biological tasks.

Our main contributions are as follows: (1) we propose a bioinformatics benchmark (Bio-benchmark) including 30 important tasks related to protein, RNA, RNA-binding protein (RBP), drug, electronic health record, medical QA and traditional Chinese medicine. We further test these tasks in zero-shot and few-shot (with CoT) settings across six mainstream vanilla LLMs without fine-tuning: GPT-4o~\cite{gpt4o}, Qwen-2.5-72b~\cite{hui2024qwen2}, Llama-3.1-70b~\cite{dubey2024llama}, Mistral-large-2~\cite{Mistral-Large-2}, Yi-1.5-34b~\cite{Yi1.5}, and InternLM-2.5-20b~\cite{cai2024internlm2}; (2) we propose an answer extraction tool (BioFinder) for LLMs on bioinformatics tasks to accurately extract answers from LLMs responses, which exceeds the accuracy of existing methods by over 40\%; (3) we evaluate and analyze the answers extracted by xFinder~\cite{yu2024xfinder} and BioFinder on different tasks, identify which benchmarked tasks are well-suited for current state-of-the-art LLMs, and propose a prompt-based approach to enhance LLM performance on tasks that are currently less effectively addressed.

\section{Benchmark construction}

\paragraph{Protein.} Datasets for predicting protein secondary structure and function are developed using data from the Protein Data Bank (PDB), involving deduplication and length restrictions to manage computational load. The protein secondary structure dataset was diversified through random sampling based on sequence lengths. For species prediction, sampling is guided by sequence distributions across species and family sequence counts, ensuring a balanced representation of families. These preprocessing steps are critical for building robust models that accurately predict protein functionalities or species, advancing research in this field.

We involve four subtasks in the protein section, including protein family sequence design, protein species prediction, protein inverse-folding design, and protein structure prediction. For instance, targeting on protein inverse-folding design task, given the protein secondary structure "CCSHHHHHHHHHHHHHHHHHHHHHHCCCCC", we prompt the language model to design the protein sequence whose secondary structure matched the given one. Protein structure prediction tasks are used to predict the secondary structure for a given specific protein sequence. For protein family sequence design, we aim to generate protein sequences belonging to a specific protein family based on the protein family ID, like "PF01713". For protein species classification, we utilize a language model to predict the category of each protein based on its sequence and the predefined protein species categories.

\paragraph{RNA.} The dataset construction for RNA secondary structure prediction, inverse folding, and functional prediction involved meticulous preprocessing across multiple sources. The bpRNA dataset provide the foundation for RNA structure and folding benchmarks, where sequences underwent deduplication, removal of entries exceeding 1024 nucleotides, and random sampling based on sequence length distributions to ensure diversity and representativeness. 

For RNA functional prediction, data from RNACentral are used, with samples select based on the distribution of sequences across different functions and families. This approach ensure a balanced representation of various RNA families and functions, facilitating the development of robust models for accurately predicting RNA characteristics and behaviors in molecular biology research.

For the RNA section, we involve 5 subtasks, including RNA family sequence design, RNA function prediction, RNA inverse-folding design, RNA structure prediction, sgRNA efficiency prediction. The first four tasks are consistent with those in the protein section, while the last involve predicting the efficiency of a given sgRNA, such as "AAAAAAAAUUGGAGGCCUUUCCCCUGGGCA", with an example efficiency prediction of 90\%.

\paragraph{RNA-binding protein.} To rigorously define a benchmark set for RNA-protein interaction-related tasks, we first obtain a set of RNA sequences experimentally verified to interact with various RNA-binding proteins. We then categorize these proteins by their binding domains, such as RNA recognition motif (RRM) and zinc finger. For the sampled protein of each binding domain, we curate balanced positive and negative RNA sequences binding to this protein. Notably, the same RNA-binding protein can demonstrate varying behavior in different species, so the species information is also included in the prompt benchmark. The classification benchmark is then constructed accordingly by prompting the model with the protein name, species, and RNA sequence information. 

In this section, we involve RNA-binding protein prediction tasks. Given a specific RNA sequence like "AGAAGGUGUGAGAUUAAUGGAUGGGGUAGCUGACG", we prompt the language model to decide whether it can bind to a specific protein like "Pp\_0237" or not. 

\paragraph{Drug.} In constructing a comprehensive drug benchmark for machine learning-guided drug design, three essential tasks are addressed: antibiotic design, drug-drug interactions, and drug-target interactions. These tasks are critical as they encompass the full spectrum of drug development, from discovery through to clinical application, and directly impact the success rate, efficiency, and safety of pharmaceuticals. For antibiotic design, the data includes novel antibiotic structures to address the urgent need for treatments against antimicrobial resistance and to test the predictive capabilities of LLMs. In the area of drug-drug interactions, the dataset is carefully curated to include 86 different interaction types, ensuring a broad representation of potential clinical scenarios to enhance medication safety and efficacy. For drug-target interactions, the selected data ensures diversity and specificity with non-overlapping drug structures and targets with low sequence similarity, streamlining the pathway from experimental screening to clinical application.

We focus on three essential tasks: drug-drug interaction prediction, drug-target interaction prediction, and drug design. For example, in the task of drug design, we prompt the model to determine if a molecule like "Cc1c(Br)c(=O)c(C(=O)N2CC(C)S(=O)C(C)C2) cn1C1CC1" can have potent efficacy on A. baumannii (a bacteria). And the example answer should be ``Not potent''.

\paragraph{Electronic health record.} EHR data is scarce, particularly within the MIMIC database~\cite{johnson2023mimic}, which predominantly contains data from ICUs and emergency departments, and is encumbered by privacy-related access restrictions. To address this, three diverse benchmarks have been selected for utilizing EHR data effectively: predicting diagnostic outcomes using patient information and test results, analyzing patient data incrementally to devise treatment plans, and generating medical reports from doctor-patient dialogues. These benchmarks are essential for improving health outcomes and the efficiency of medical services.

\paragraph{General medical question-answering.} The selected data for medical knowledge benchmarks emphasize accuracy and are derived from expert-verified exam questions. These benchmarks are geographically diverse: MedQA sources from exams in Mainland China, Taiwan, and the USA; MedMCQA from Indian exams; and HeadQA from Spanish exams. They cover a wide range of topics, with MMCU including 13 subcategories like medical imaging and infectious diseases, while HeadQA spans six areas such as medicine and psychology. Some datasets simulate real-world medical consultations requiring long-text responses to open-ended questions, although most benchmarks use multiple-choice formats due to limited alternatives, reflecting a compromise in developing practical, 0-shot style medical question answering.

\paragraph{Questiong-Answering in traditional Chinese medicine.} High-quality benchmarks for TCM are rare, primarily consisting of multiple-choice formats that often mix Eastern and Western medical concepts, making it difficult to isolate pure TCM content. The CMB/CMMLU datasets fill this gap by incorporating questions from ancient texts and traditional TCM, expressed in Classical Chinese, which tests models' understanding of historical language nuances. Furthermore, the TCM SD dataset uses real clinical TCM cases, where questions focus on predicting diseases and syndromes from patient information using specific TCM terminologies, ensuring a practical evaluation of models' clinical capabilities in TCM.

\begin{figure*}[t]
    \centering
    \includegraphics[width=16cm]{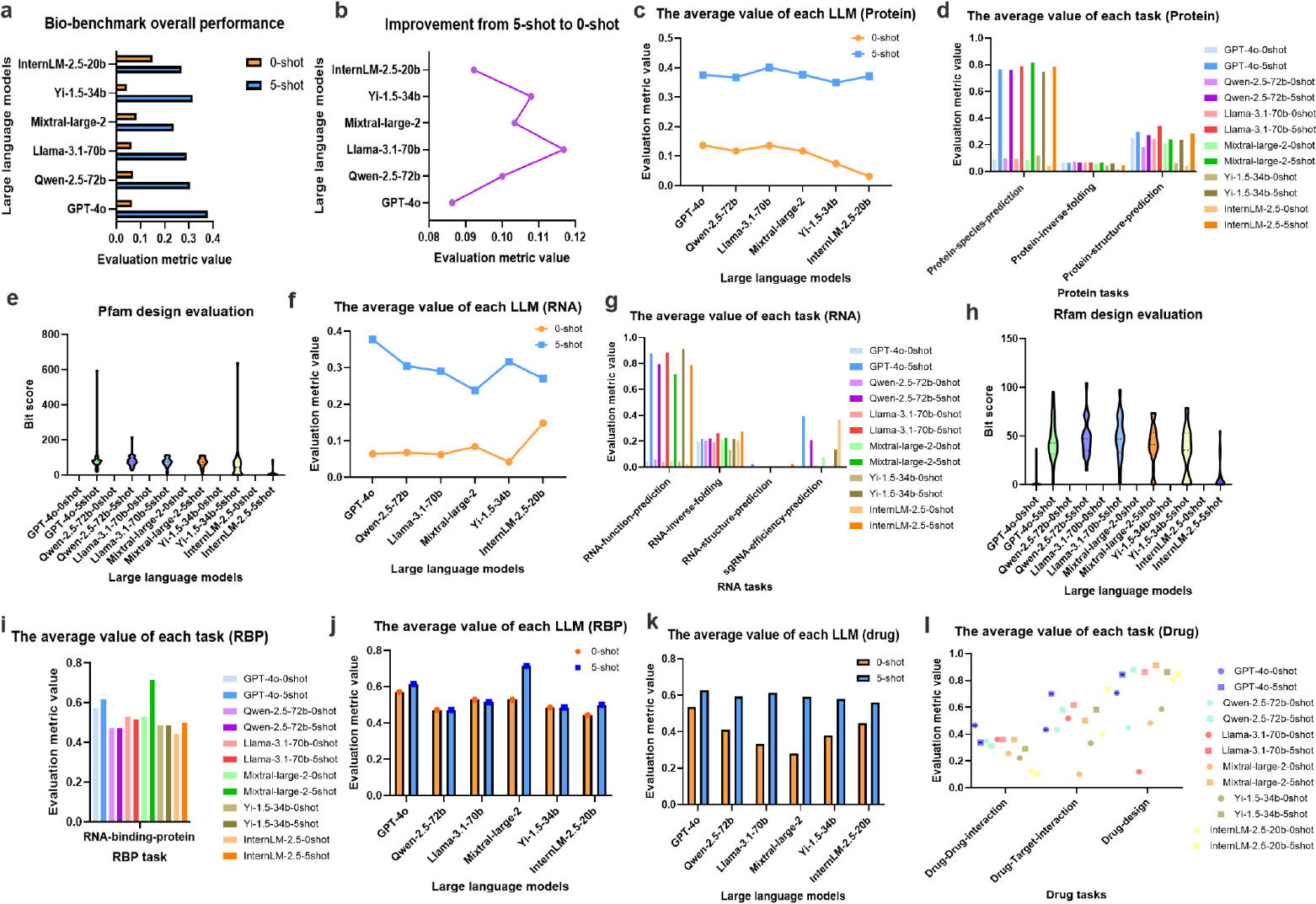}
    \caption{The performance of sequence and drug benchmarks in the Bio-benchmark. \textbf{a, b} represent overall performance. \textbf{c, d, e} indicate performance on protein benchmark. \textbf{f, g, h} show performance on RNA benchmark. \textbf{i, j} reflect performance on RBP benchmark. \textbf{k, l} demonstrate performance on drug benchmark.}
    \label{bb1}
\end{figure*}

\section{Evaluation methods}

\subsection{Problem definition}
\label{ProblemDefine}

Given a set of questions $Q = \{ q_i \mid q_i \in \Sigma^* \}$, our goal is to evaluate the performance of LLMs on various tasks. Since we adopt an unconstrained output format using the CoT method, accurately extracting the standard answer $a_i$ from the generated text $y_i = \text{LLM}(q_i)$ is challenging; traditional RegEx-based methods struggle to balance false positives and negatives. To address this, we utilize the BioFinder framework (Section~\ref{bioexp1}) with an answer extraction function $E: \Sigma^* \rightarrow \Sigma^*$ to extract the key answer $k_i = E(y_i)$ from the model output $y_i$.

We divide the evaluation tasks into objective and subjective evaluations. For \textbf{objective evaluation tasks} ($t_i \in T_{\text{obj}}$), such as multiple-choice questions and character matching with a definitive standard answer $a_i$, we define an evaluation function $M_{\text{obj}}: \Sigma^* \times \Sigma^* \times T_{\text{obj}} \rightarrow \mathbb{R}$ to compare the extracted answer $k_i$ with the standard answer $a_i$ and compute the performance of LLM on task $t_i$:
\begin{equation}
\text{Performance}_{\text{obj}} = M_{\text{obj}}(k_i, a_i, t_i).
\end{equation}

For \textbf{subjective evaluation tasks} ($t_i \in T_{\text{subj}}$), such as long-text generation, open-ended question answering, and case analysis, we evaluate the quality of the model's output through the following:

\begin{itemize}[leftmargin=*,label={},itemsep=0ex,topsep=0ex,partopsep=0ex,parsep=0ex]
\item\textbf{Similarity calculation}: Define a similarity function $S: \Sigma^* \times \Sigma^* \rightarrow \mathbb{R}$ to compute the textual and semantic similarity between the generated answer $y_i$ and the reference answer $a_i$:
\begin{equation}
s_i = S(y_i, a_i).
\end{equation}

\item\textbf{Expertise assessment}: Use advanced LLMs like GPT-4o to assess the professional quality of the generated content $y_i$, obtaining a quality score $q_i$.

\item\textbf{Logical consistency judgment}: Use the trained BioFinder to perform Natural Language Inference (NLI), defining a logical relation function:
\begin{equation}
\begin{split}
N: \Sigma^* \times \Sigma^* &\rightarrow \\ 
&\{ \text{Entailment}, \text{Contradiction}, \text{Neutral} \}
\end{split}
\end{equation}

to conduct fine-grained analysis of the logical relationship between $y_i$ and $a_i$:
\begin{equation}
n_i = N(y_i, a_i).
\end{equation}
\end{itemize}

Through the BioFinder framework, we can efficiently and accurately extract key answers from LLM-generated texts without constraining the output format, enabling a comprehensive evaluation of the model's capabilities and generation quality.

\subsection{Fine-grained textual evaluation}

To evaluate the quality of the text generation, we employ metrics including similarity, expertise, and logical consistency. Specifically, inspired by FACTSOCRE and OLAPH~\cite{min2023factscore,jeong2024olaph}, we compare the LLM's response $\hat{P}$ with reference statements categorized as \textit{Must-Have} (MH) and \textit{Nice-to-Have} (NH), assigned weights $w_{\text{MH}}$ and $w_{\text{NH}}$, respectively, with $w_{\text{MH}} > w_{\text{NH}}$. Using the BioFinder framework and \textbf{Natural Language Inference} (NLI), we classify the relationship between the LLM's response and each reference statement as Entailed, Contradicted, or Neutral. Based on this classification, we define the following metrics:

\textbf{Comprehensiveness} measures the extent to which the model's response correctly includes the reference statements, defined as:
\begin{equation}
\text{Comprehensiveness}(\hat{P}) = \frac{\sum_{x \in \text{Entailed}} w(x)}{\sum_{x \in S} w(x)},
\end{equation}
where $S = \text{MH} \cup \text{NH}$ represents the set of all reference statements, $\text{Entailed}$ is the subset of statements classified as entailed by BioFinder, and $w(x)$ is the weight of statement $x$.

\textbf{Hallucination Rate} evaluates the proportion of contradictory information in the LLM's response relative to the reference statements, reflecting the degree of misinformation, defined as:
\begin{equation}
\text{Hallucination rate}(\hat{P}) = \frac{\sum_{x \in \text{Contradicted}} w(x)}{\sum_{x \in S} w(x)},
\end{equation}
where $\text{Contradicted}$ is the subset of statements classified as contradicted by BioFinder.

\textbf{Omission rate} measures the extent to which the model's response omits reference statements (especially must-have information), defined as:
\begin{equation}
\text{Omission Rate}(\hat{P}) = \frac{\sum_{x \in \text{Neutral}} w(x)}{\sum_{x \in S} w(x)},
\end{equation}
where $\text{Neutral}$ is the subset of statements classified as neutral by BioFinder.

To assess the stability of the LLM's performance across the dataset, we calculate the variability of the metrics and define \textbf{Consistency} as:
\begin{equation}
\text{Consistency} = 1 - \frac{\sigma_{\text{Comp}} + \sigma_{\text{Hall}} + \sigma_{\text{Omit}}}{\sum{\sigma_{\text{max}}}},
\end{equation}
where $\sigma_{\text{Comp}}$, $\sigma_{\text{Hall}}$, $\sigma_{\text{Omit}}$ are the standard deviations of Comprehensiveness, Hallucination Rate, and Omission Rate, respectively, and $\sigma_{\text{max}}$ is the sum of maximum possible standard deviations.

We combine the above metrics to define the model's overall performance score: 
\begin{align}
\text{Overall Score} &= \alpha \overline{\text{Comp}} - \beta \overline{\text{Hall}} \notag \\
&\quad - \gamma \overline{\text{Omit}} + \delta \text{Consistency},
\end{align}
where $\overline{\text{Comp}}$, $\overline{\text{Hall}}$, $\overline{\text{Omit}}$ are the average values of Comprehensiveness, Hallucination Rate, and Omission Rate, respectively, and $\alpha$, $\beta$, $\gamma$, $\delta$ are weighting coefficients satisfying $\alpha + \beta + \gamma + \delta = 1$.

\begin{figure*}[t]
    \centering
    \includegraphics[width=16cm]{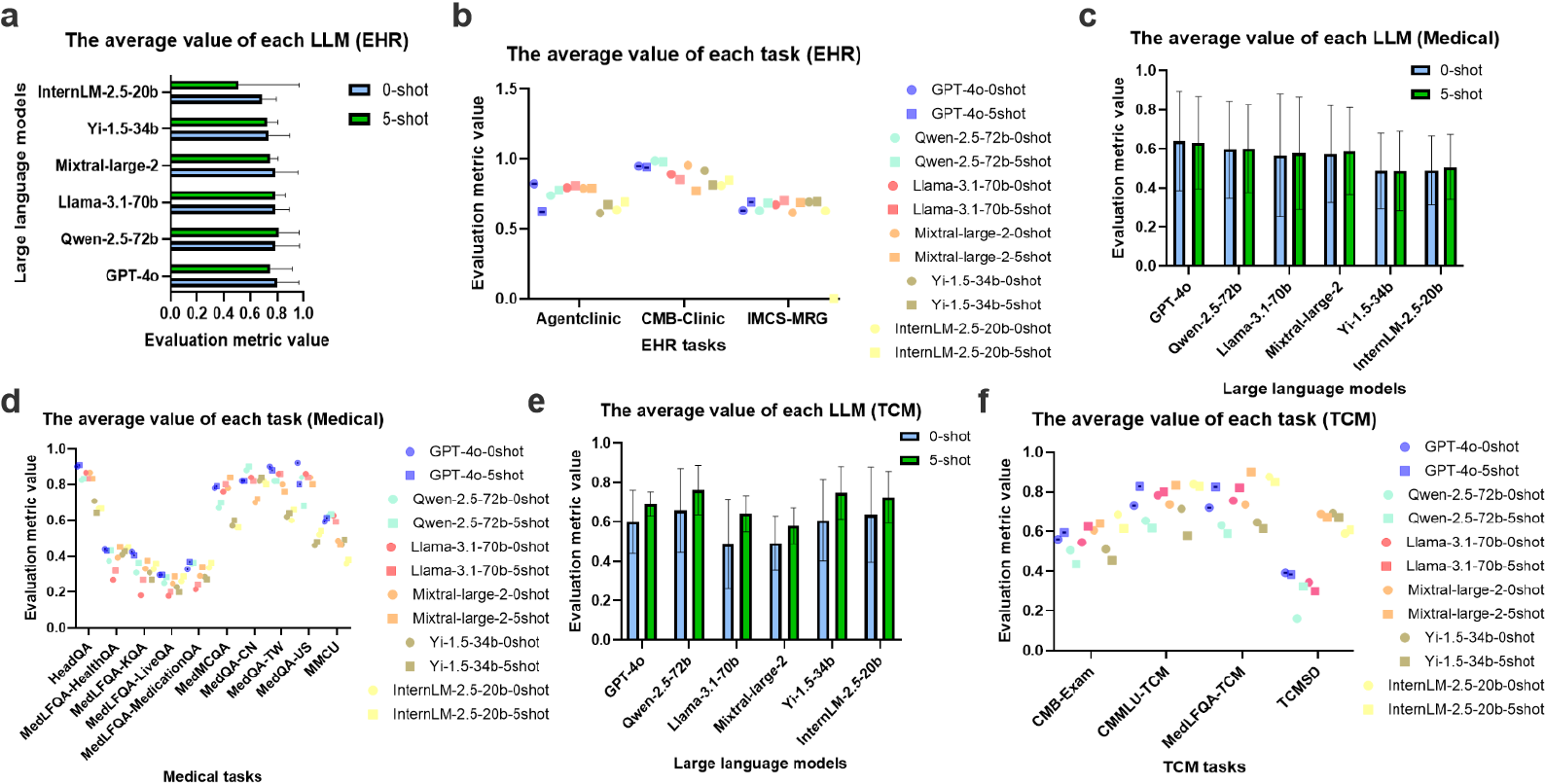}
    \caption{The performance of biotext benchmarks in the Bio-benchmark. \textbf{a, b} represent performance on EHR benchmark. \textbf{c, d} indicate performance on Medical-QA benchmark. \textbf{e, f} show performance on TCM benchmark.}
    \label{bb2}
\end{figure*}

\section{Experiment and results}


\subsection{BioFinder}
\label{bioexp1}
\subsubsection{Experiment Setting}
As outlined in Section \ref{ProblemDefine}, evaluating LLMs presents unique challenges. Although the xFinder framework demonstrated optimal performance on diverse tasks like multiple-choice questions, short text matching, and mathematical extraction, it fell short on the Bio-benchmark dataset \cite{yu2024xfinder}. To bridge this gap, we adapt xFinder-llama38it\footnote{\url{https://huggingface.co/IAAR-Shanghai/xFinder-llama38it}} into BioFinder, enhancing its capability for biological sequence extraction and long-text Natural Language Inference (NLI) tasks without reference answers.

For the biological sequence extraction, we manually annotate 1,428 samples from six models in Bio-benchmark, focusing on 0-shot and five-shot outputs, and iteratively retrained using GPT4o and RegEx on misextracted sequences \cite{2023opencompass}. The NLI task involve collecting 13,698 samples, applying multi-round reasoning with GPT4o, and refining outputs through self-consistency checks and manual reviews \cite{wang2022self}.

Training uses advanced fine-tuning methodologies such as LoRA, PRoLoRA, QLoRA, and MoS \cite{hu2021loralowrankadaptationlarge, wang2024lora, wang2024prolora, wang2024mos, dettmers2024qlora}, specifically employing the Xtuner framework with QLoRA. Prompts are strategically designed for each task to enhance the model's adherence to instructions.

\subsubsection{Baselines and Evaluation}
\paragraph{Objective Answer Extraction}
For objective tasks such as multiple-choice questions, short text matching, and mathematical answer extraction, we fine-tune xFinder to enhance its bio-sequence extraction capabilities while retaining all its original functionalities. Table~\ref{complete} compares the performance of BioFinder, regular expressions, and GPT-4. BioFinder achieved a 93.5\% extraction accuracy in bio-sequence extraction tasks, exceeding regular expressions by over 30\%, and delivering twice the extraction quality of GPT-4 using the same prompt. Across all objective tasks, BioFinder's average extraction accuracy reached 94.7\%, comparable to single-round human review, outperforming regular expressions at 72.1\% and GPT-4 at 62.9\%.




\paragraph{Subjective Task Evaluation}

We evaluate subjective tasks by comparing BioFinder with GPT-4\footnote{\url{https://openai.com/index/gpt-4/}}, RoBERTa-large-mnli\footnote{\url{https://huggingface.co/FacebookAI/roberta-large-mnli}}, and GTE-large-en-v1.5\footnote{\url{https://huggingface.co/Alibaba-NLP/gte-large-en-v1.5}}. The comparison involve mapping model outputs and reference answers into a shared embedding space and assessing cosine similarity. BioFinder outperform the others on a medical NLI test, achieving 89.8\% accuracy compared to 59.9\% for GPT-4, 27.0\% for RoBERTa-large-mnli, and 23.4\% for GTE-large-en-v1.5, demonstrating state-of-the-art performance.

RoBERTa-large-mnli, not initially design for long sequences ($>$512 tokens), required text segmentation and a bagging approach to mitigate semantic interruptions. In addition, embedding models are found to be ineffective for NLI tasks, often misclassifying antonym sentences as similar, contrary to their expected 'Contradiction' classification in NLI tasks, as detailed in Appendix \ref{EmbeddingModelCaseStudy}.

\subsection{Bio-benchmark}
\label{setbio}
\subsubsection{Experiment Setting}

We conduct experiments on the Bio-benchmark datasets. We use Openai-API and 8*A100-80G GPUs to infer LLMs with LM-Deploy and Huggingface backend~\cite{2023lmdeploy, huggingface_inference}. We employ sampling hyperparameters with top-p set to 1.0 and a temperature of 0.2 for generation (Specific prompts in the Appendix~\ref{promptstem}). We use BioFinder to extract the answers of objective evaluation and execute NLI tasks.

\subsubsection{Evaluation metrics}
\label{metric}
To evaluate the performance of Bio-benchmark tasks on LLMs, we use the following metrics: (1) Bit score~\cite{ye2006blast, nawrocki2009infernal, nawrocki2013infernal}: Measures the alignment quality of sequences, normalizing the raw alignment score by database size and randomness. (2) Recovery rate: Indicates the proportion of relevant instances accurately retrieved from the total relevant instances available. (3) Accuracy (Acc.): Evaluates the proportion of correct predictions made out of the total predictions. (4) \textbf{BERTScore}~\cite{zhang2019bertscore}: We adopt BERTScore to compare the similarity between embeddings of a generated sentence and the reference sentence. (5) Rouge-L: Evaluates the similarity of text based on the longest common subsequence, focusing on sequence order and content overlap in summaries or translations. (6) GPT evaluation metrics (fluency, relevance, completeness, medicine proficiency).


\begin{table*}[h]
\label{BioFinderTable}
\small
\centering
\begin{tabular}{l|c|c|c|c|c}
\toprule
\textbf{} & \textbf{RegEx} & \textbf{GPT-4o} & \textbf{BioFinder} & \textbf{RoBERTa-large-mnli} & \textbf{GTE-large-en-v1.5}\\
\midrule
MCQ Matching   & 77.5 & 65.8 & \textbf{95.5}&  -   &  -   \\
Text Matching              & 74.8 & 80.5 & \textbf{94.3}&  -   &  -   \\
Numerical Matching         & 68.1 & 67.0 & \textbf{95.5}&  -   &  -   \\
\textbf{Sequence Extraction}        & 68.0 & 38.5 & \textbf{93.5}&  -   &  -   \\
\textbf{Overall Objective Tasks}    & 72.1 & 62.9 & \textbf{94.7}&  -   &  -   \\
\textbf{NLI Tasks}  &   -  & 59.9 & \textbf{89.8}& 27.0 & 23.4 \\
\bottomrule
\end{tabular}
\caption{Comparison of performance on different tasks between RegEx (OpenCompass), GPT-4o, RoBERTa-large-mnli, GTE-large-en-v1.5, and BioFinder.}
\end{table*}





  
\subsection{Evaluation metrics}
\label{metric}
For more detailed Bio-benchmark experimental evaluation metrics, please see Appendix~\ref{metric}.

\textbf{1. Protein:} Pfam design (bit score), protein species prediction (Acc.), protein inverse folding (recovery rate), protein structure prediction (recovery rate). \textbf{2. RNA:} RNA function prediction (Acc.), RNA inverse folding (recovery rate), RNA structure prediction (F1), Rfam design (bit score), sgRNA efficiency prediction (Acc.). \textbf{3. RBP:} RNA-binding protein (Acc.). \textbf{4. Drug:} Drug-Drug interaction (Acc.), drug-target interaction (Acc.), drug design (Acc.). \textbf{5. EHR:} Agentclinic (Acc.), CMB-Clinic (fluency, relevance, completeness, medical proficiency), IMCS-MRG (Rouge-L). \textbf{6. Medical:} HeadQA (Acc.), MedLFQA-HealthQA (COMPREHENSIVENESS), MedLFQA-KQA (COMPREHENSIVENESS), MedLFQA-LiveQA (COMPREHENSIVENESS), MedLFQA-MedicationQA (COMPREHENSIVENESS), MedMCQA (Acc.), MedQA-CN (Acc.), MedQA-TW (Acc.), MedQA-US (Acc.), MMCU (Acc.). \textbf{7. TCM:} CMB-Exam (Acc.), CMMLU-TCM (Acc.), MedLFQA-TCM (Acc.), TCMSD (Acc.).

\section{Analysis}
We observe that prompt formats significantly affect LLM performance in biological sequence inference. Specifically, continuous bio-sequences versus those separated by spaces or newline characters lead to different tokenizations due to LLMs using Byte Pair Encoding (BPE). BPE tokenizers often treat several consecutive uppercase letters as a single token, which impairs the ability of LLMs to understand and generate bio-sequences. Figure \ref{fig:protein} showed that separating sequences with newline characters achieves over three times the alignment accuracy compared to continuous inputs.

In addition, from Figure~\ref{bb1}, Figure~\ref{bb2} and Table~\ref{complete}, we can obtain the following results.

\subsection{Protein benchmark}

In the protein species prediction, few-shot significantly improved accuracy across all LLMs. Yi-1.5-34b experience at least a sixfold increase, while InternLM-2.5-20b witness a maximum nearly twentyfold increase. Mistral-large-2 lead with 82\% accuracy, followed by Llama-3.1-70b at 79\%.

For inverse protein folding and protein structure prediction, we measure the sequence recovery rate. The latter task show notably better performance, with accuracy approximately four times higher than the former. Notably, in a five-shot setting, Llama-3.1 70b achieve the highest score of 34\% in protein structure prediction.

In addition, we evaluate the proficiency of LLMs in generating protein sequences from specific Pfam IDs. Initially, all LLMs score zero in 0-shot tests, indicating their inability to generate accurate sequences based solely on Pfam IDs. However, with 10-shot prompting, LLMs substantially improve, achieving bit scores over 50\%, with GPT-4o reaching the highest at 87\%.

\subsection{RNA benchmark}

 We evaluate large language models on RNA sequence processing tasks. For RNA function prediction, similar to protein species prediction, few-shot (5-shot) prompting significantly enhance model performance over 0-shot, with Llama-3.1-70b 5-shot achieving the highest accuracy of 89\%.

In RNA-inverse folding and structure prediction tasks, results are generally disappointing. However, RNA inverse folding outperforms structure prediction tasks, with Llama-3.1-70b 5-shot leading with a recovery rate of 26\%. Few-shot prompting notably improves outcomes in both RNA tasks.

We also examine sgRNA efficiency prediction, crucial for CRISPR gene editing success. GPT-4o 5-shot and InternLM-2.5-20b 0-shot both score above 30\%. Notably, InternLM-2.5-20b performed better in 0-shot than in 5-shot, where it scores zero, suggesting no benefit from extra examples.

Lastly, we evaluate RNA sequence generation from specific Rfam IDs, using bit scores to judge sequence quality. 10-shot prompting consistently outperforms 0-shot across all LLMs, though the average score of 40.78\% is lower than the 72.63\% achieved in the protein section under the same conditions.

\subsection{RBP Benchmark}
The RNA binding protein task is used to evaluate the ability of LLMs to predict whether a given RNA sequence can bind to a specific protein. These LLMs achieved an average accuracy of 53\%, with the highest scores being 71\% by Mistral-large-2 (5-shot) and 61\% by GPT-4o (5-shot). In this task, the few-shot prompting strategy slightly improve performance compared to 0-shot prompting.

\subsection{Drug Benchmark}
According to Figure \ref{fig:drug}, we conduct experiments on drug sequences similar to those performed on biological sequences. By counting the number of atoms in the drug sequences, we discover that the prompt format significantly impacts the LLMs' drug-sequence generation. In the drug design task, large language models are tested on their ability to predict the efficacy of molecular entities (SMILES strings) against specific bacteria. Using 5-shot prompting, all LLMs perform well, achieving accuracy scores above 80\%, with Mistral-large-2 5-shot topping at 91\%. Notably, 5-shot prompting boost the performance of Llama-3.1-70b significantly, from 12\% in 0-shot to 86\%.

In the Drug-Drug and Drug-Target interaction tasks, LLMs evaluate potential interactions. Drug-Drug interaction predictions are unsatisfactory across all models, with the best score being 0.47 by GPT-4o in 0-shot, which decreases to 0.34 with 5-shot prompting. Conversely, Drug-Target interaction predictions show better results; InternLM-2.5-20b 5-shot and GPT-4o 5-shot achieve the highest scores of 73\% and 70\%, respectively. 5-shot prompting significantly enhances performance, increasing the average accuracy from 62\% in 0-shot to 37\%.

\subsection{EHR Benchmark}

In the EHR benchmark, LLMs show high accuracy across three medical reasoning subtasks: AgentClinic, CMB-Clinic, and IMCS-MRG. For the AgentClinic task, average metric values are 73.2\% (0-shot) and 72.7\% (5-shot), with GPT-4o achieving a high of 82.24\% in 0-shot.

Performance in the CMB-Clinic task was excellent, with average metric values of 0.917 in 0-shot and 0.867 in five-shot settings using CoT reasoning. This performance closely matches reference answers, highlighting the potential of LLMs in medical case diagnosis.

It is noted that for the CMB-Clinic task, the accuracy of GPT-4o slightly decreased in the five-shot setting comparing with the 0-shot setting. This decrease is attributed to interference from example questions and the inclusion of irrelevant information in few-shot prompting, potentially leading to reduced accuracy when longer inputs do not improve reasoning outcomes.

\subsection{Medical Benchmark}

For the Medical-QA Benchmark, which includes seven datasets, the average performance of LLMs in 0-shot and 5-shot settings is similar. However, the overall performance varied significantly across different datasets. LLMs perform well on multiple-choice problems including HeadQA, MedMCQA, MedQA-CN, MedQA-TW, and MedQA-USA, with average 0-shot metric values of 80.6\%, 68.7\%, 81.3\%, 76.7\%, and 74.0\%, respectively. Medical QA tasks are particularly well-suited for large language models, which align effectively with the domain. 


Furthermore, there is little to no improvement with five-shot prompting, and in some cases, performance decreased. The reason can be that biomedical QA is a task that aligns well with LLMs, achieving high 0-shot performance. The additional five-shot examples provide minimal benefit.

\subsection{TCM Benchmark}
Overall, LLMs perform well on the TCM benchmark. Out of the four subtasks, only TCMSD show relatively lower performance with an average metric value of 31.7\%. Notably, there is a significant improvement in five-shot compared to 0-shot settings across various subtasks. The average metric value improves from 31.7\% to 65.3\%. This is because TCM is less commonly considered during model training, and the use of prompts can enhance performance. We infer that if LLMs are fine-tuned with the related corpus, the performance could improve further.

\section{Conclusion}

This study successfully establish a comprehensive benchmarking framework for evaluating the performance of various LLMs on bioinformatics tasks.  Utilizing the BioFinder tool, we are able to accurately extract key answers from LLM outputs, significantly enhancing the accuracy of answer extraction.  Our results demonstrate that LLMs perform well in multiple subdomains of bioinformatics, such as protein, RNA, and drug design tasks, particularly in few-shot learning settings.  Future research may further optimize prompt engineering strategies to enhance the efficiency and precision of LLMs on specific tasks.

\section*{Limitations}
Despite the comprehensive nature of the Bio-benchmark and the novel approach of the BioFinder tool, our study has several limitations that warrant consideration. Firstly, the benchmark primarily assesses the intrinsic capabilities of LLMs using 0-shot and few-shot settings without fine-tuning. While this approach provides insights into the raw abilities of these models, it may not fully reflect their performance in practical, real-world applications where fine-tuning on specific tasks can significantly enhance effectiveness.
 
Secondly, while we cover a broad range of biological tasks, the selection of these tasks and their corresponding datasets might still not be representative of all possible bioinformatics challenges. The diversity and complexity of biological data are such that even a comprehensive benchmark like ours cannot encapsulate all potential tasks and scenarios. This limitation could affect the generalizability of our findings and recommendations.

Furthermore, the performance of the BioFinder tool, although superior to existing methods by a significant margin, is heavily dependent on the quality and structure of the data fed into the LLMs. In scenarios where the LLM outputs are ambiguous or overly complex, the extraction accuracy of BioFinder might be compromised. This limitation underscores the ongoing challenge in the field of bioinformatics to develop methods that are robust across a wide variety of data types and quality.

Additionally, our benchmark does not address the computational and financial costs associated with deploying large language models. The use of such models, particularly in 0-shot and few-shot scenarios, can be resource-intensive. This aspect could limit the accessibility of our proposed methods for researchers with limited resources.

Lastly, while our study indicates promising areas for the application of LLMs in bioinformatics, it also reveals that certain tasks remain challenging for these models. Developing LLM architectures or training strategies that can tackle these hard-to-model aspects of biological data remains a key area for future research. The ongoing development of LLMs should focus on enhancing adaptability and accuracy in these less tractable areas to broaden the utility of LLMs in bioinformatics.
\section*{Ethics Statement}

This paper does not involve an ethics statement.

\bibliography{custom}

\appendix

\section{Appendix}
\label{sec:appendix}

\subsection{Results}
\begin{table*}[htbp!]
    \centering
    \tiny 
    \setlength{\tabcolsep}{0.5pt}
    \renewcommand{\arraystretch}{1.3} 
    \begin{tabularx}{\textwidth}{l l r *{12}{>{\centering\arraybackslash}X}}
        \toprule
        
        \textbf{Task} & \textbf{Subtask} & \textbf{Count} & \multicolumn{2}{c}{\textbf{GPT-4o}} & \multicolumn{2}{c}{\textbf{InternLM-2.5 20b}} & \multicolumn{2}{c}{\textbf{Llama-3.1 70b}} & \multicolumn{2}{c}{\textbf{Mistral-large-2}} & \multicolumn{2}{c}{\textbf{Qwen 2.5-72b}} & \multicolumn{2}{c}{\textbf{Yi-1.5 34b}} \\

        & & & 0-shot & 5-shot & 0-shot & 5-shot & 0-shot & 5-shot & 0-shot & 5-shot & 0-shot & 5-shot & 0-shot & 5-shot \\
        \midrule
        \textbf{Protein} & Protein-species-prediction & 200 & 9.00 & 76.50 & 4.00 & 78.50 & 9.50 & 79.00 & 8.50 & 82.00 & 10.00 & 76.00 & 12.00 & 75.00 \\
        \textbf{Protein} & Protein-inverse-folding & 264 & 6.97 & 6.29 & 1.46 & 4.64 & 6.69 & 6.88 & 5.65 & 6.70 & 7.02 & 6.95 & 4.48 & 5.77 \\
        \textbf{Protein} & Protein-structure-prediction & 264 & 25.09 & 29.79 & 3.96 & 28.21 & 24.63 & 34.31 & 21.02 & 24.23 & 18.11 & 27.13 & 5.99 & 23.99 \\

        \textbf{RBP} & RNA-binding-protein & 70 & 57.14 & 61.43 & 44.29 & 50.00 & 52.86 & 51.43 & 52.86 & 71.43 & 47.14 & 47.14 & 48.57 & 48.57 \\
        \textbf{RNA} & RNA-function-prediction & 280 & 4.64 & 87.86 & 2.14 & 78.57 & 3.93 & 88.57 & 4.64 & 71.79 & 6.07 & 79.29 & 3.93 & 91.07 \\
        \textbf{RNA} & RNA-inverse-folding & 200 & 19.76 & 21.71 & 20.64 & 27.19 & 19.50 & 26.25 & 20.62 & 22.90 & 20.28 & 21.92 & 12.99 & 21.82 \\
        \textbf{RNA} & RNA-structure-prediction & 200 & 0.50 & 2.14 & 0.00 & 2.26 & 0.07 & 0.75 & 0.65 & 0.10 & 0.43 & 0.07 & 0.09 & 0.37 \\
        \textbf{RNA} & sgRNA-efficiency-prediction & 300 & 0.67 & 39.33 & 36.67 & 0.00 & 1.33 & 0.67 & 7.67 & 0.33 & 0.33 & 20.67 & 0.00 & 13.33 \\

        \textbf{Drug} & Drug-Drug-interaction & 86 & 46.51 & 33.72 & 12.79 & 10.47 & 36.05 & 36.05 & 25.58 & 36.05 & 34.88 & 31.40 & 22.09 & 29.07 \\
        \textbf{Drug} & Drug-Target-interaction & 60 & 43.33 & 70.00 & 40.00 & 73.33 & 51.67 & 61.67 & 10.00 & 50.00 & 43.33 & 58.33 & 33.33 & 58.33 \\
        \textbf{Drug} & Drug-design & 58 & 70.69 & 84.48 & 81.03 & 84.48 & 12.07 & 86.21 & 48.28 & 91.38 & 44.83 & 87.93 & 58.62 & 86.21 \\
        
        \textbf{EHR} & Agentclinic & 214 & 82.24 & 62.15 & 63.55 & 69.16 & 79.44 & 80.84 & 78.97 & 78.97 & 73.83 & 77.57 & 61.21 & 67.29 \\
        \textbf{EHR} & CMB-Clinic & 74 & 94.93 & 93.92 & 80.95 & 84.66 & 88.90 & 85.14 & 95.54 & 77.08 & 98.56 & 97.97 & 91.55 & 81.35 \\
        \textbf{EHR} & IMCS-MRG & 200 & 63.02 & 69.02 & 62.82 & 0.00 & 67.21 & 70.21 & 61.56 & 68.64 & 62.98 & 68.48 & 69.20 & 69.40 \\

        \textbf{Medical} & HeadQA & 120 & 90.00 & 90.83 & 66.67 & 66.67 & 86.67 & 83.33 & 86.67 & 83.33 & 82.50 & 83.33 & 70.83 & 64.17 \\
        \textbf{Medical} & MedLFQA-HealthQA & 50 & 43.76 & 43.03 & 44.27 & 45.03 & 26.75 & 31.84 & 39.19 & 45.24 & 37.37 & 43.20 & 40.81 & 42.67 \\
        \textbf{Medical} & MedLFQA-KQA & 50 & 42.34 & 40.41 & 32.73 & 35.89 & 18.12 & 26.76 & 32.93 & 37.55 & 30.82 & 36.25 & 30.77 & 26.88 \\
        \textbf{Medical} & MedLFQA-LiveQA & 50 & 29.57 & 29.60 & 26.05 & 28.69 & 17.76 & 19.95 & 24.80 & 28.67 & 25.07 & 28.25 & 22.82 & 19.85 \\
        \textbf{Medical} & MedLFQA-MedicationQA & 50 & 32.70 & 36.94 & 33.54 & 36.31 & 21.45 & 24.25 & 28.99 & 33.84 & 35.82 & 36.26 & 28.51 & 27.08 \\
        \textbf{Medical} & MedMCQA & 100 & 78.00 & 79.00 & 56.00 & 56.00 & 76.00 & 80.00 & 78.00 & 84.00 & 67.00 & 70.00 & 57.00 & 60.00 \\
        \textbf{Medical} & MedQA-CN & 50 & 82.00 & 82.00 & 82.00 & 80.00 & 84.00 & 82.00 & 70.00 & 72.00 & 88.00 & 90.00 & 82.00 & 84.00 \\
        \textbf{Medical} & MedQA-TW & 50 & 90.00 & 88.00 & 60.00 & 66.00 & 86.00 & 86.00 & 80.00 & 76.00 & 82.00 & 82.00 & 62.00 & 64.00 \\
        \textbf{Medical} & MedQA-US & 50 & 92.00 & 80.00 & 52.00 & 54.00 & 86.00 & 84.00 & 84.00 & 80.00 & 84.00 & 68.00 & 46.00 & 48.00 \\
        \textbf{Medical} & MMCU & 142 & 59.15 & 61.27 & 35.92 & 38.03 & 62.68 & 59.15 & 48.59 & 46.48 & 62.68 & 63.38 & 46.48 & 49.30 \\

        \textbf{TCM} & CMB-Exam & 200 & 56.00 & 60.50 & 62.50 & 61.50 & 50.50 & 51.00 & 43.50 & 45.50 & 59.50 & 64.00 & 54.50 & 68.50 \\
        \textbf{TCM} & CMMLU-TCM & 185 & 72.97 & 73.51 & 80.00 & 82.70 & 65.41 & 71.35 & 61.62 & 57.84 & 82.70 & 83.24 & 78.38 & 83.78 \\
        \textbf{TCM} & MedLFQA-TCM & 200 & 72.00 & 73.50 & 82.00 & 85.00 & 63.00 & 64.50 & 59.00 & 61.50 & 82.50 & 90.00 & 75.50 & 87.50 \\
        \textbf{TCM} & TCMSD & 200 & 39.00 & 68.75 & 30.00 & 60.75 & 16.00 & 69.25 & 32.25 & 67.00 & 38.25 & 67.25 & 34.50 & 59.00 \\

        \bottomrule
    \end{tabularx}
    \caption{Performance of Various LLMs on Bioinformatics Tasks}
    \label{complete}
\end{table*}

\begin{figure}[t]
    \centering
    \includegraphics[width=4cm]{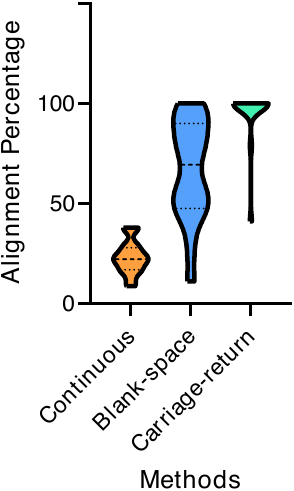}
    \caption{The proportion of Alignment by different methods.}
    \label{fig:protein}
\end{figure}

\begin{figure*}[t]
    \centering
    \includegraphics[width=16cm]{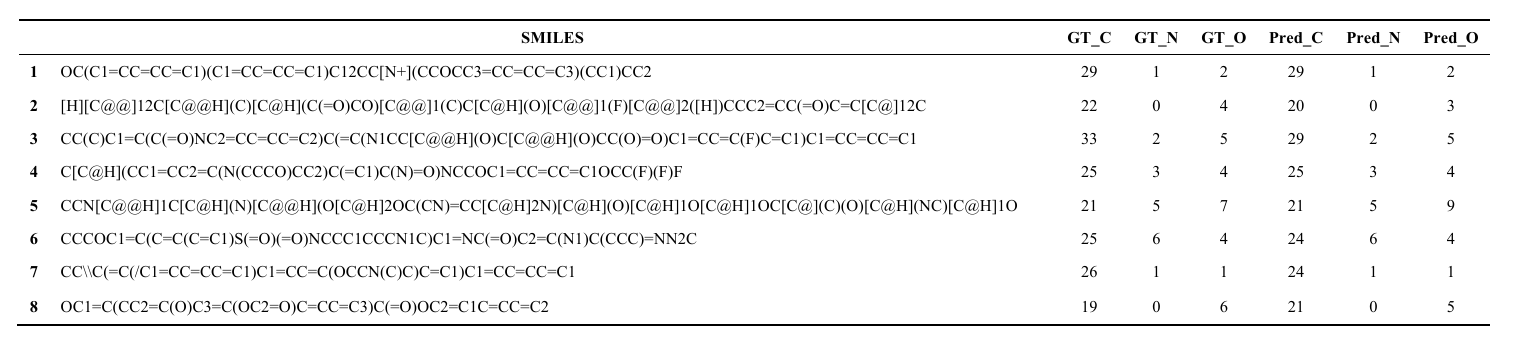}
    \caption{The proportion of Alignment by different methods.}
    \label{fig:drug}
\end{figure*}

\subsection{Related Work}



\subsubsection{Bioinformatics Benchmark}
With the emergence of LLMs for various biological tasks, there is also a growing number of benchmark methods designed for different inputs, including RNA, protein, and biomedicine-related texts. 
Protein is one of the first biological fields that LLM has been employed with multiple established benchmarks, including protein fitness prediction~\cite{notin2023proteingym}, protein design~\cite{ye2024proteinbench}, and multi-task ones~\cite{xu2022peer}.
As for the more recently studied RNA LLMs, benchmarks have been created to include various structure-, function-, and engineering-related tasks~\cite{runge2024rnabench,ren2024beacon}.
For LLMs working with biomedical textual data, there have also been various benchmark datasets, focusing on EHR~\cite{bae2024ehrxqa,soni2023quehry}, medicine QA~\cite{jin2021disease,pal2022medmcqa,jin2019pubmedqa}, and specifically traditional Chinese medicine QA~\cite{tan2023medchatzh,li2023huatuo26m}.

\subsubsection{Answer Quality Assessment}

Bio-benchmark dataset encompasses diverse tasks with complex answer formats, including objective tasks like Multiple-Choice Questions (MCQA)~\cite{robinson2022leveraging,he2024ultraeval}, short text matching, mathematical inference, and biological sequence prediction, as well as subjective tasks such as long text generation and open-ended QA. Consequently, the answer formats generated by LLMs are highly complex and varied.

LLMs' outputs are often uncontrollable; the output of different LLMs could differ significantly even with the same prompt~\cite{yukun2024improving,gu2022robustness}. Traditional evaluation frameworks attempt to constrain outputs through prompts, but this can negatively impact generation accuracy and reasoning ability~\cite{rottger2024political}. For complex tasks, LLMs' instruction-following ability decreases, leading to answers that do not adhere to specific formats~\cite{asai2023self}.


Traditional evaluation frameworks like LM Eval Harness and OpenCompass typically use regular expressions (RegEx) or Judge models to extract answers~\cite{gao2024raw,contributors2023opencompass,dubois2024length,zheng2023judging}, aiming to cover various output formats. This approach requires significant effort in designing RegEx patterns and struggles to balance false positives and omissions~\cite{zhu2023judgelm,wang2023pandalm}. Moreover, the generality and accuracy of judge models are significantly inferior to advanced LLMs or human evaluation~\cite{huang2024empirical}, and the cost of using advanced LLMs or human evaluators on large datasets is prohibitively high. To address these challenges, xFinder proposed a framework that uses LLMs for inference, achieving excellent results in answer extraction~\cite{yu2024xfinder}. However, this framework's capability and applicability in bioinformatics need enhancement; it cannot extract answers from data without reference answers and currently cannot assess the quality of long-text responses.


\subsubsection{Bioinformatics Prompting}

\paragraph{Biological Sequences} 
In bioinformatics, prompting techniques have been applied to both protein and nucleic acid sequences. 
For DNA sequences, template-based prompts generate natural language explanations of gene interactions, benefiting various aspects like the interpretability of synthetic lethality predictions for drug discovery~\cite{petroni2019language,liu2021pre,zhang2024prompt}. 
Soft prompts integrate DNA sequences into tunable templates~\cite{li2023plpmpro}, improving the performance of models like DNABERT~\cite{ji2021dnabert} in recognizing promoter sequences. 
For RNA sequences, CoT and repeated prompt have been applied on RNA-seq-based cell type annotation with GPT-3 and GPT-4~\cite{hou2024assessing}.
For protein sequences, interaction prompts~\cite{wu2024guided} and continuous prompts~\cite{li2021prefix,zou2023linker} aid in predicting protein structures and interactions by embedding task-specific information directly into the input sequences.
Additionally, protein CoT has also been applied in predicting nonphysical protein-protein interaction~\cite{wang2023instructprotein}.

\paragraph{Drugs}
Prompting methods are also utilized in drug-related tasks, including drug-target binding affinity prediction: dynamic prompt generation captures unique interactions between drugs and their targets by integrating context-specific prompts with molecular features~\cite{xiao2024hgtdp}. 
Additionally, latent vector prompts enhance molecular design by incorporating continuous prompts into transformer models through cross-attention mechanisms~\cite{kong2024dual}. 
In-context learning strategies, such as those used for predicting synergistic drug combinations, leverage masking and graph representations to improve personalized drug synergy predictions~\cite{edwards2023synergpt}.

\paragraph{Biological Textual Data} 
Prompting techniques extend to biological textual data, including EHR and biomedical QA systems. 
In EHRs, explanatory prompts streamline clinical documentation by contextualizing each section, reducing the clinicians' workload~\cite{he2024shimo}. 
Prompt engineering methods, including paraphrasing and persona-based instructions, enhance the effectiveness of LLM embeddings for medical diagnostics and prognostics~\cite{gao2024raw,chen2023extensive}. 
In biomedical QA systems, strategies like CoT~\cite{chen2024evaluating,wu2024regulogpt} and Graph of Thoughts (GoT)~\cite{hamed2024accelerating} can help improve reasoning abilities and reduce misinformation in responses. 
Specialized prompts for TCM QA systems incorporate domain-specific knowledge and examples, optimizing model performance in specialized medical contexts~\cite{yizhen2024exploring}.

\section{Appendix B}

\label{EmbeddingModelCaseStudy}

We evaluated the cosine similarities among various sentences embedded using the GTE-large-en-v1.5 model. Specifically, SENTENCE1-CONCLUSION represents the LLM's response, ANSWER denotes the reference answer, and INVERSE ANSWER is the negation of the reference answer. As illustrated in Figure~\ref{fig_ss
}, when the reference answer is negated, the similarity between the reference answer and its inverse is second only to the similarity between the CONCLUSION and the reference answer. This observation indicates that the embedding model does not effectively capture the logical and semantic relationships between sentences.

\begin{figure*}[t] \centering \includegraphics[width=14cm]{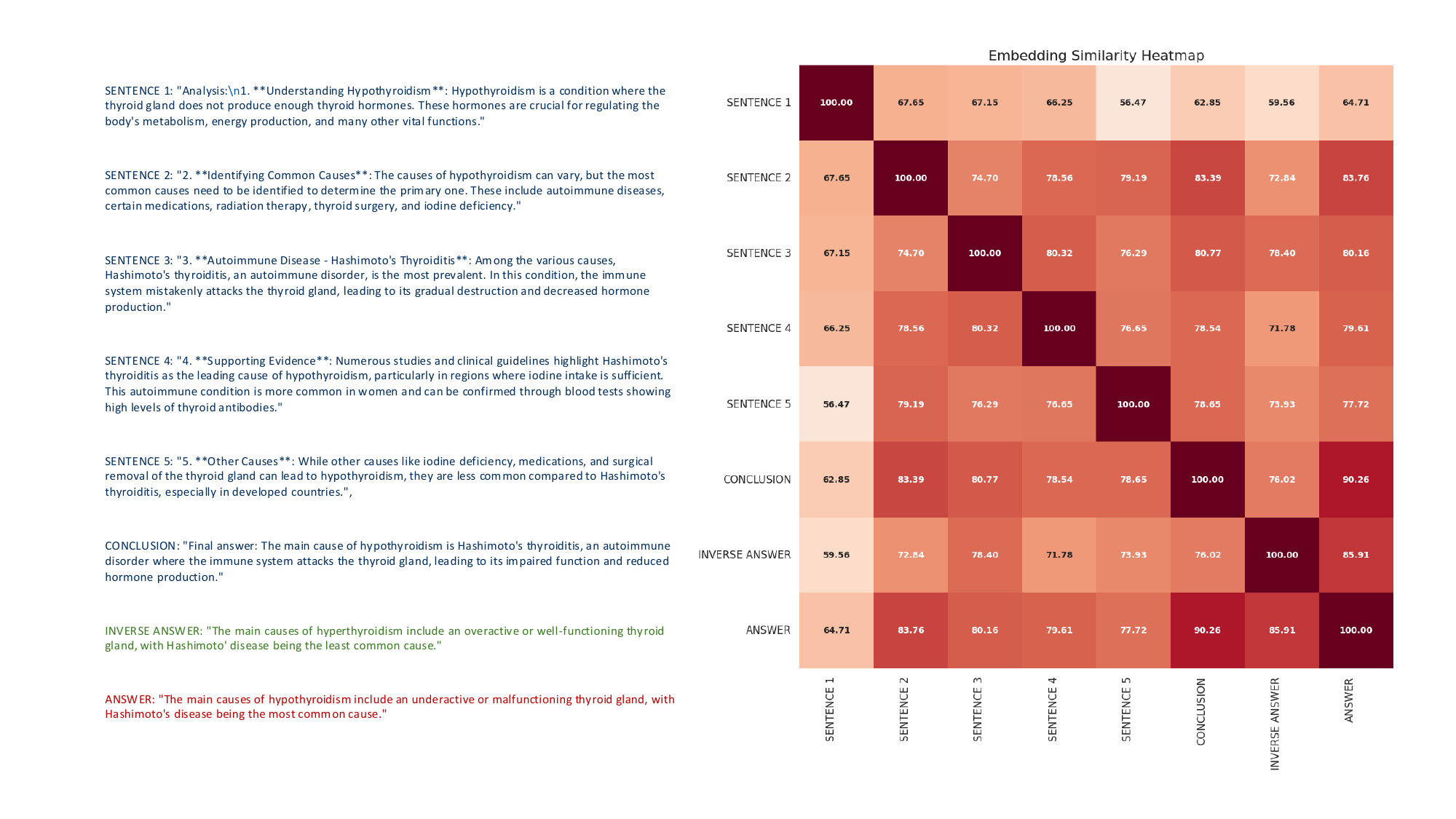} \caption{Cosine similarities between sentence embeddings using the GTE-large-en-v1.5 model. SENTENCE1-CONCLUSION is the LLM's response, ANSWER is the reference answer, and INVERSE ANSWER is the negation of the reference answer. The high similarity between the reference answer and its inverse suggests limitations in the embedding model's ability to distinguish logical negations.} \label{fig_ss
} \end{figure*}

\newpage

\section{Overall evaluation metric values}  
\begin{figure*}[t]
    \centering
    \includegraphics[width=14cm]{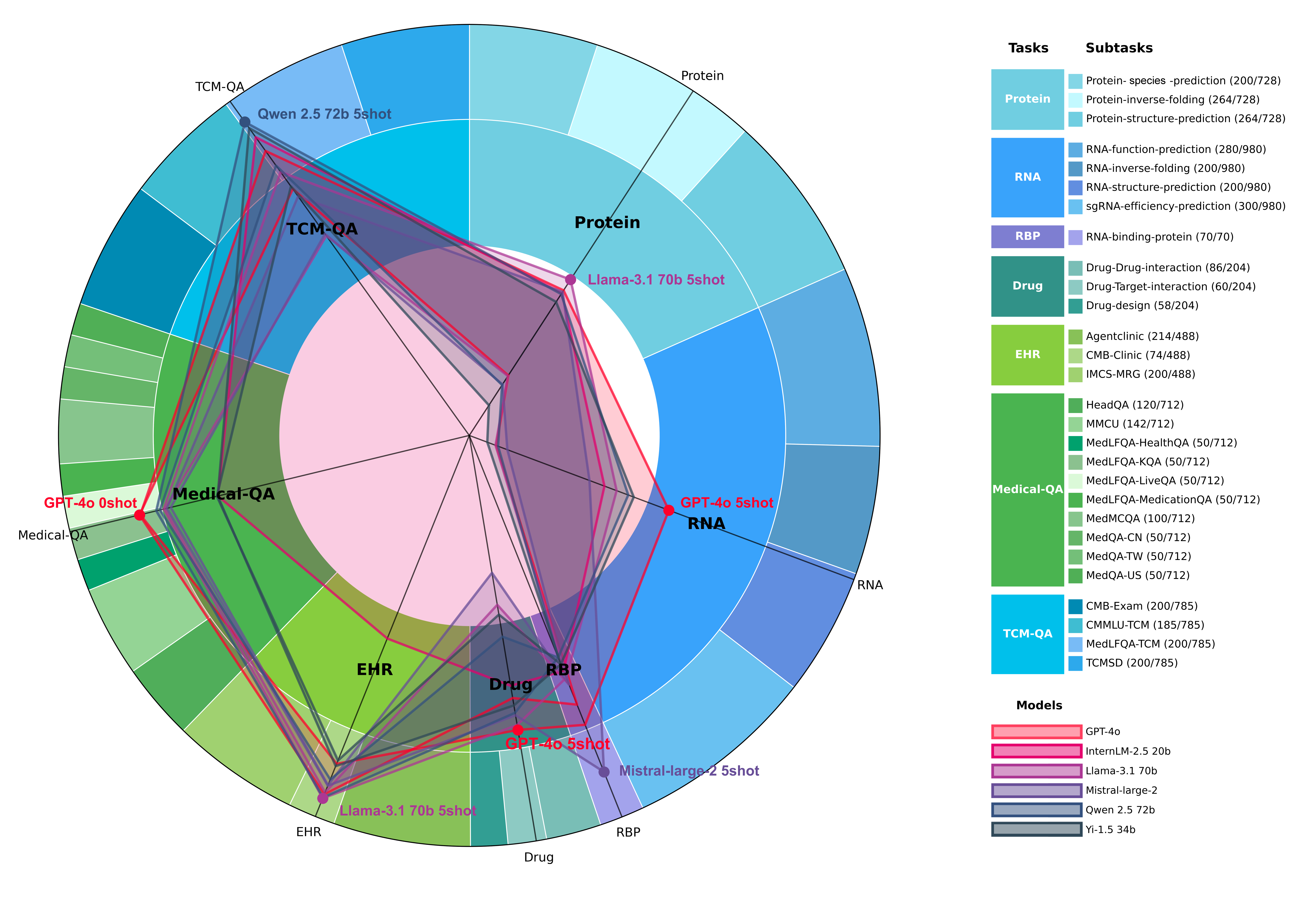}
    \caption{Overall evaluation metric values}
    \label{res}
\end{figure*}
\subsection{Bioinformatics Sequence}

\newpage

\section{Prompt Template in Inference}
\label{promptstem}
\onecolumn
\begin{figure}[h!]
    \centering
    \includegraphics[width=1\textwidth]{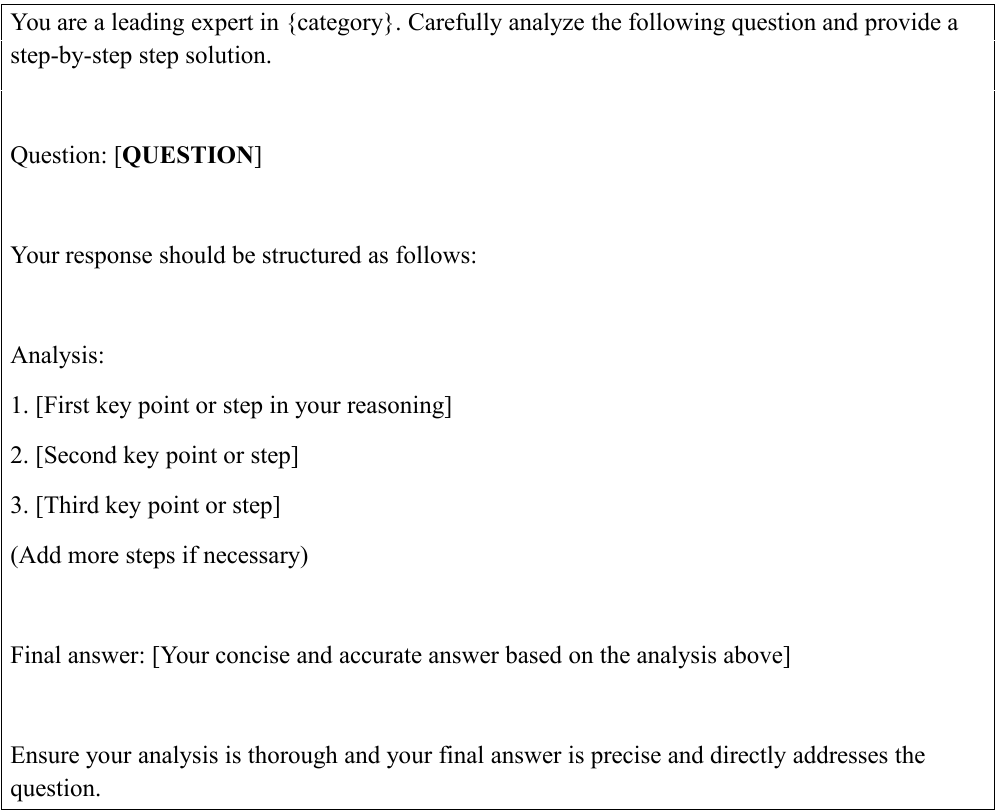} 
\end{figure}
\begin{figure}[h!]
    \centering
    \includegraphics[width=1\textwidth]{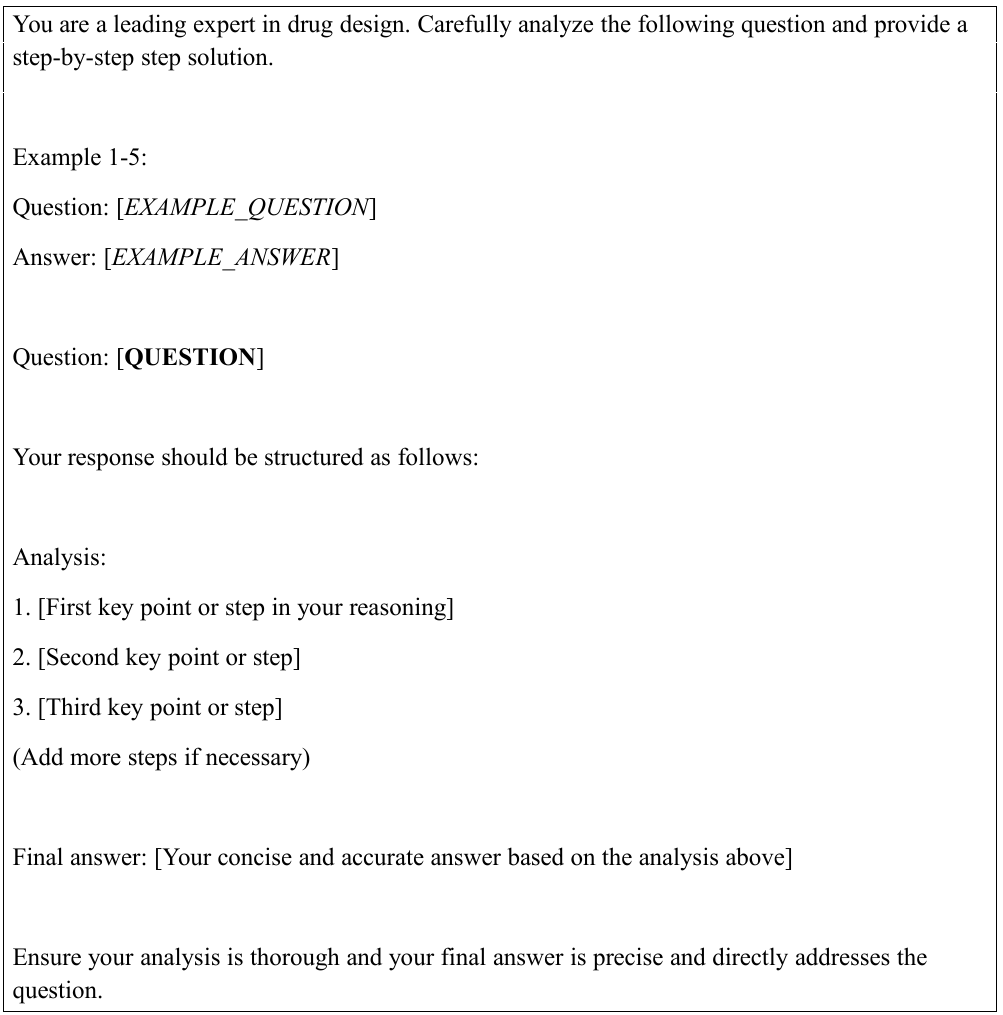} 
\end{figure}
\begin{figure}[h!]
    \centering
    \includegraphics[width=1\textwidth]{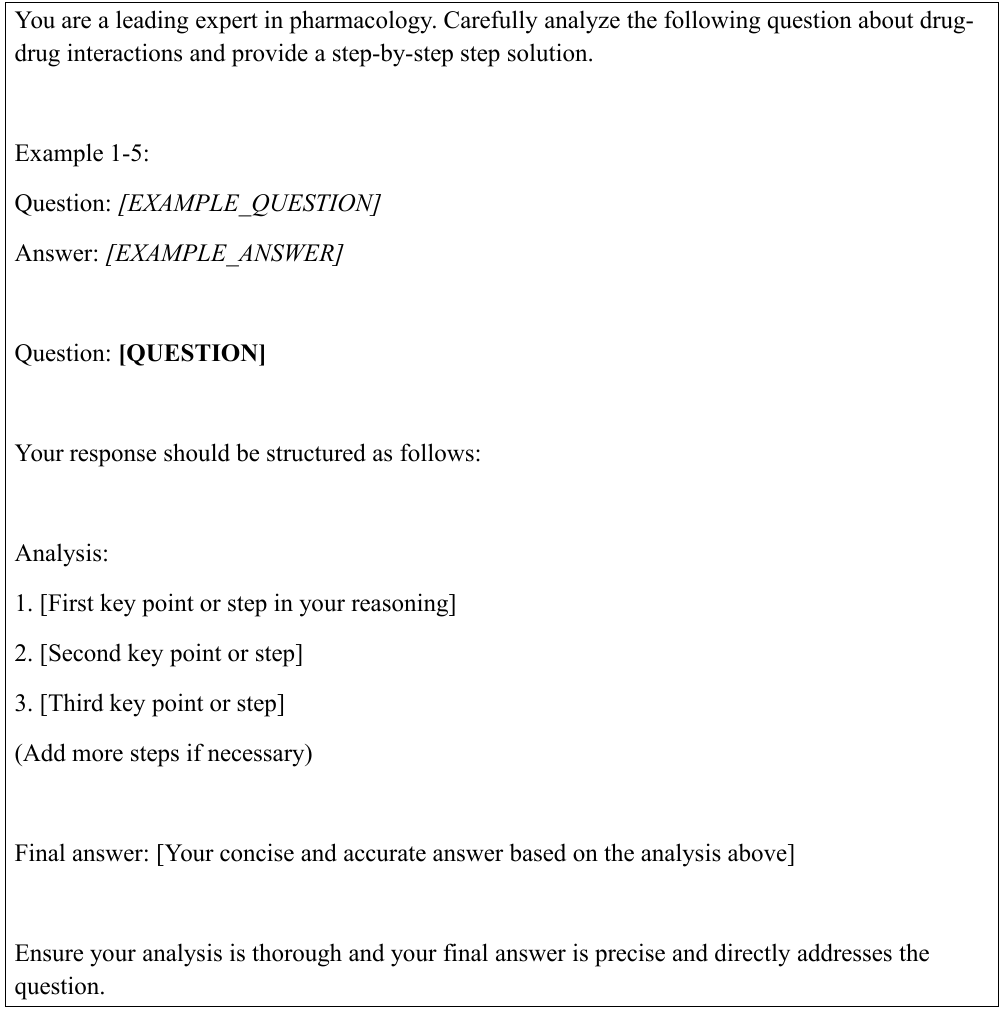} 
\end{figure}
\begin{figure}[h!]
    \centering
    \includegraphics[width=1\textwidth]{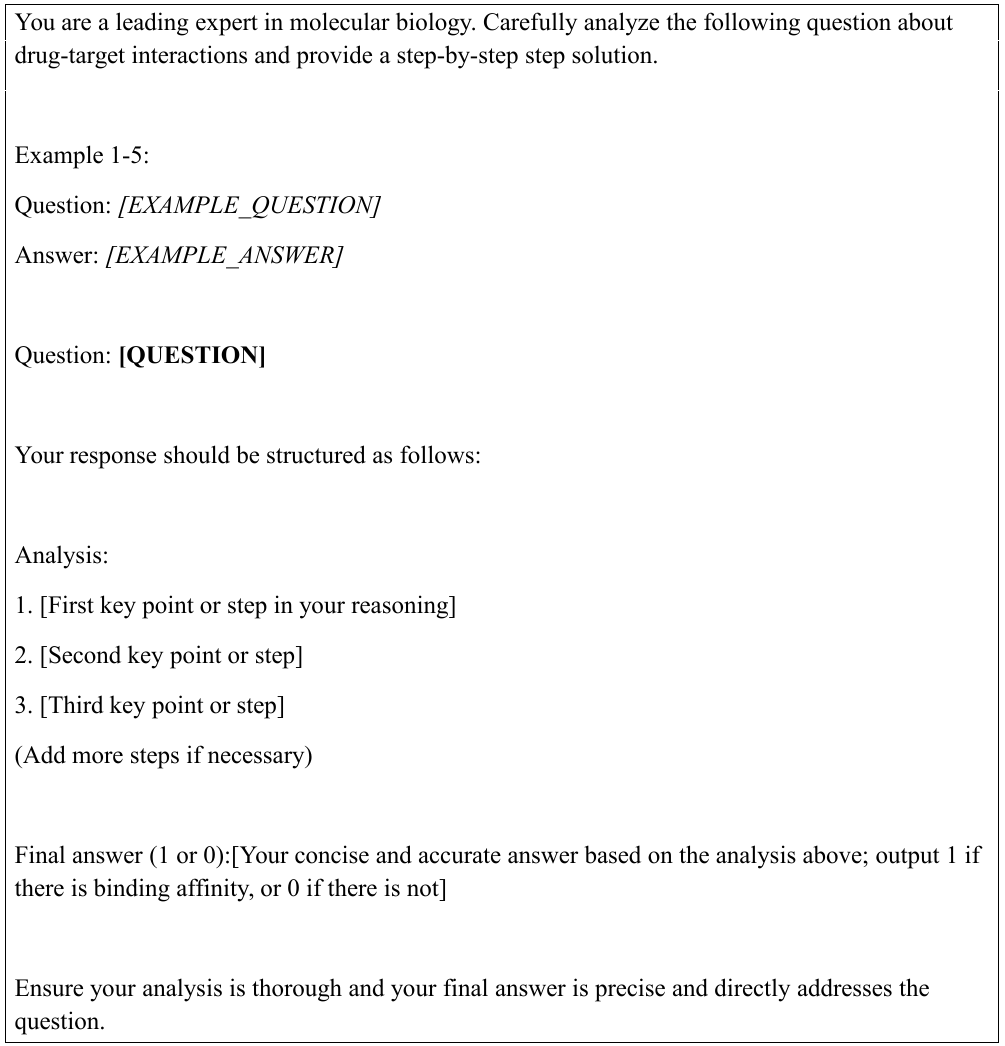} 
\end{figure}
\begin{figure}[h!]
    \centering
    \includegraphics[width=1\textwidth]{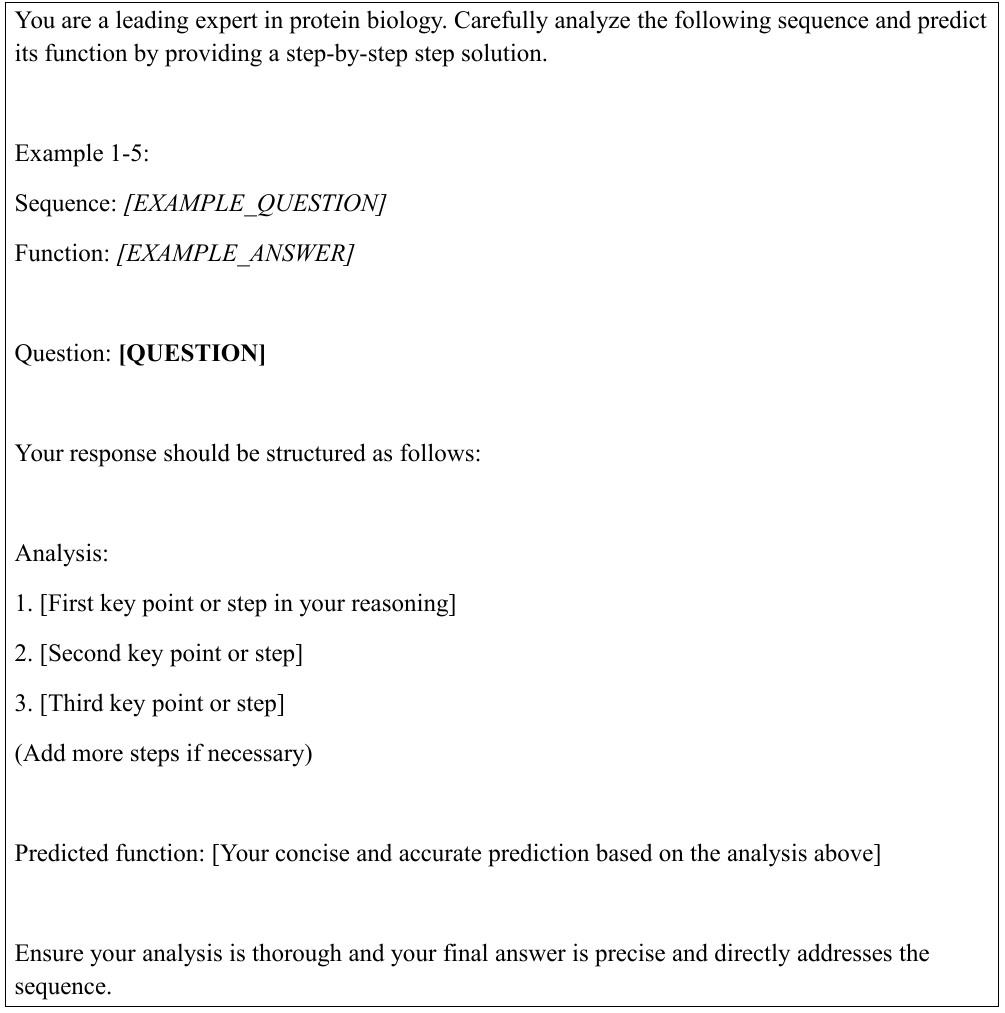} 
\end{figure}
\begin{figure}[h!]
    \centering
    \includegraphics[width=1\textwidth]{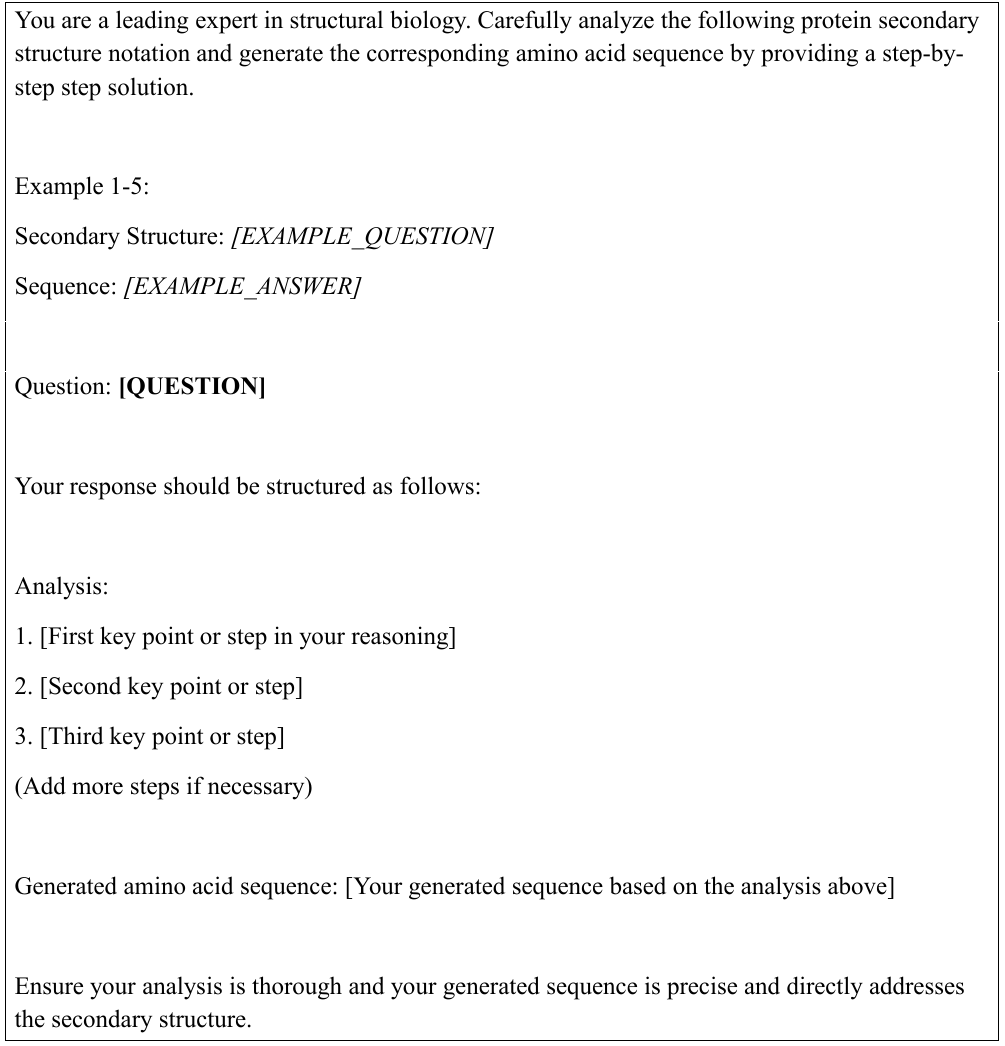} 
\end{figure}
\begin{figure}[h!]
    \centering
    \includegraphics[width=1\textwidth]{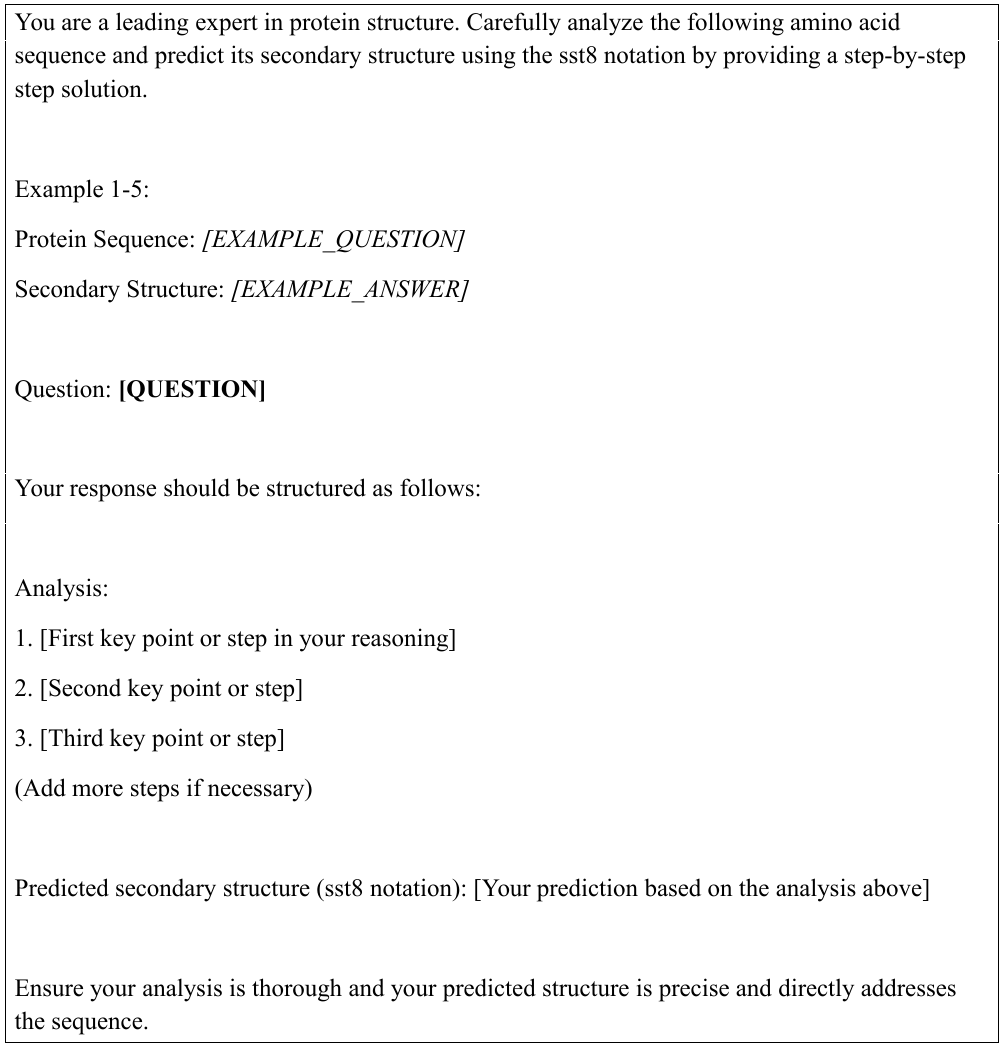} 
\end{figure}
\begin{figure}[h!]
    \centering
    \includegraphics[width=1\textwidth]{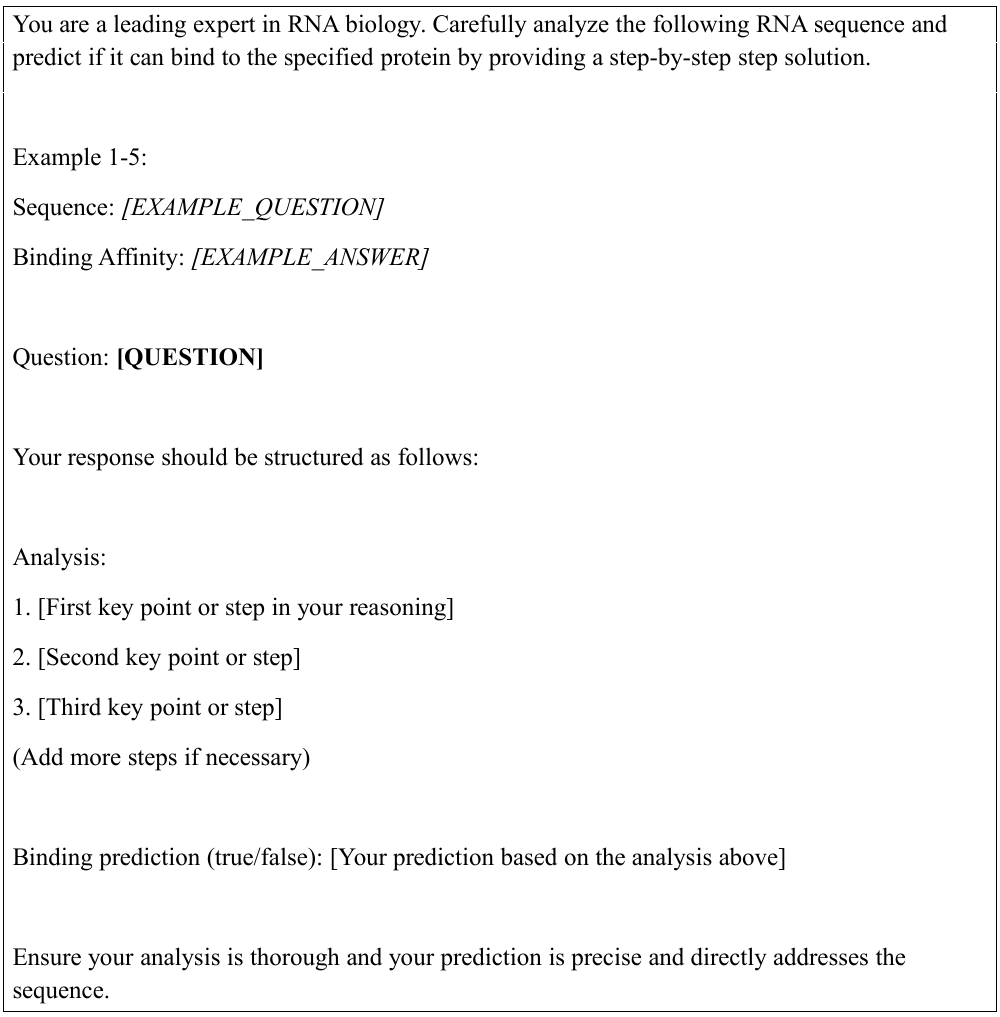} 
\end{figure}
\begin{figure}[h!]
    \centering
    \includegraphics[width=1\textwidth]{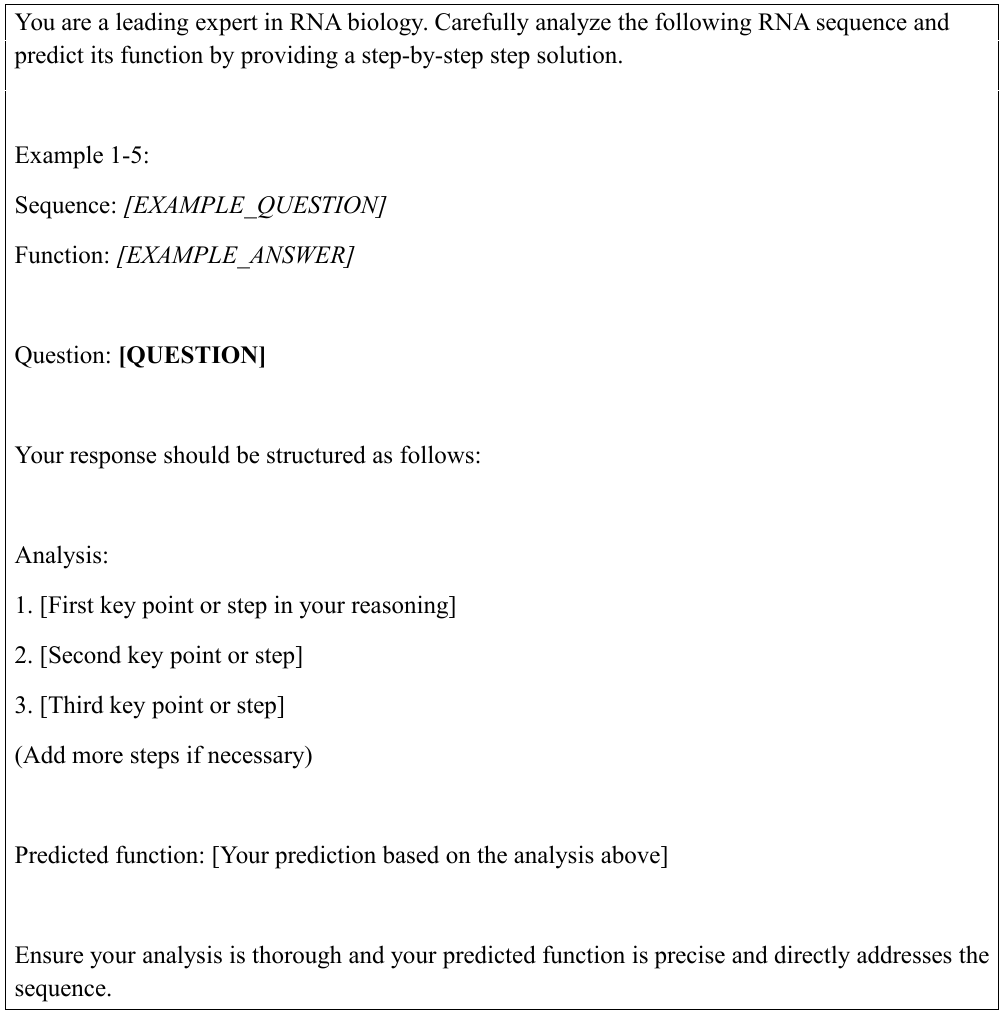} 
\end{figure}
\begin{figure}[h!]
    \centering
    \includegraphics[width=1\textwidth]{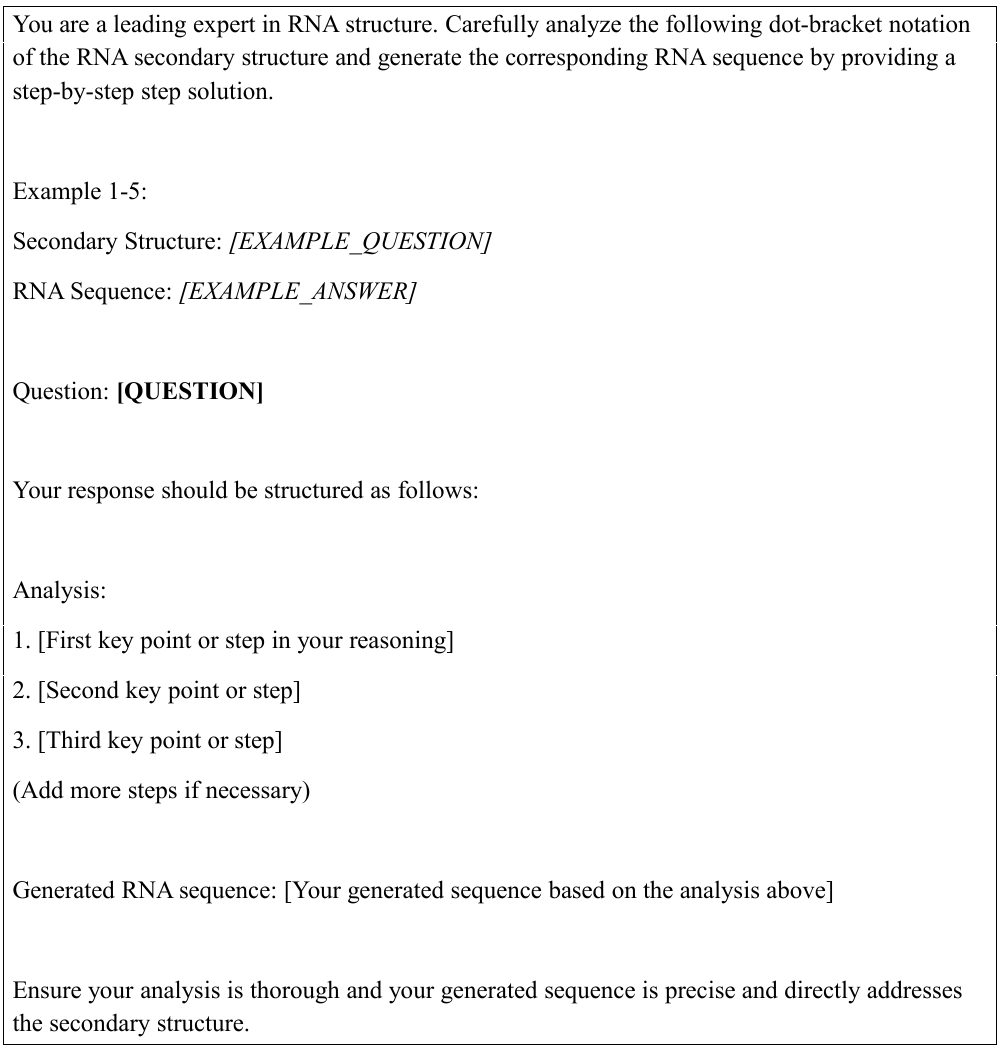} 
\end{figure}
\begin{figure}[h!]
    \centering
    \includegraphics[width=1\textwidth]{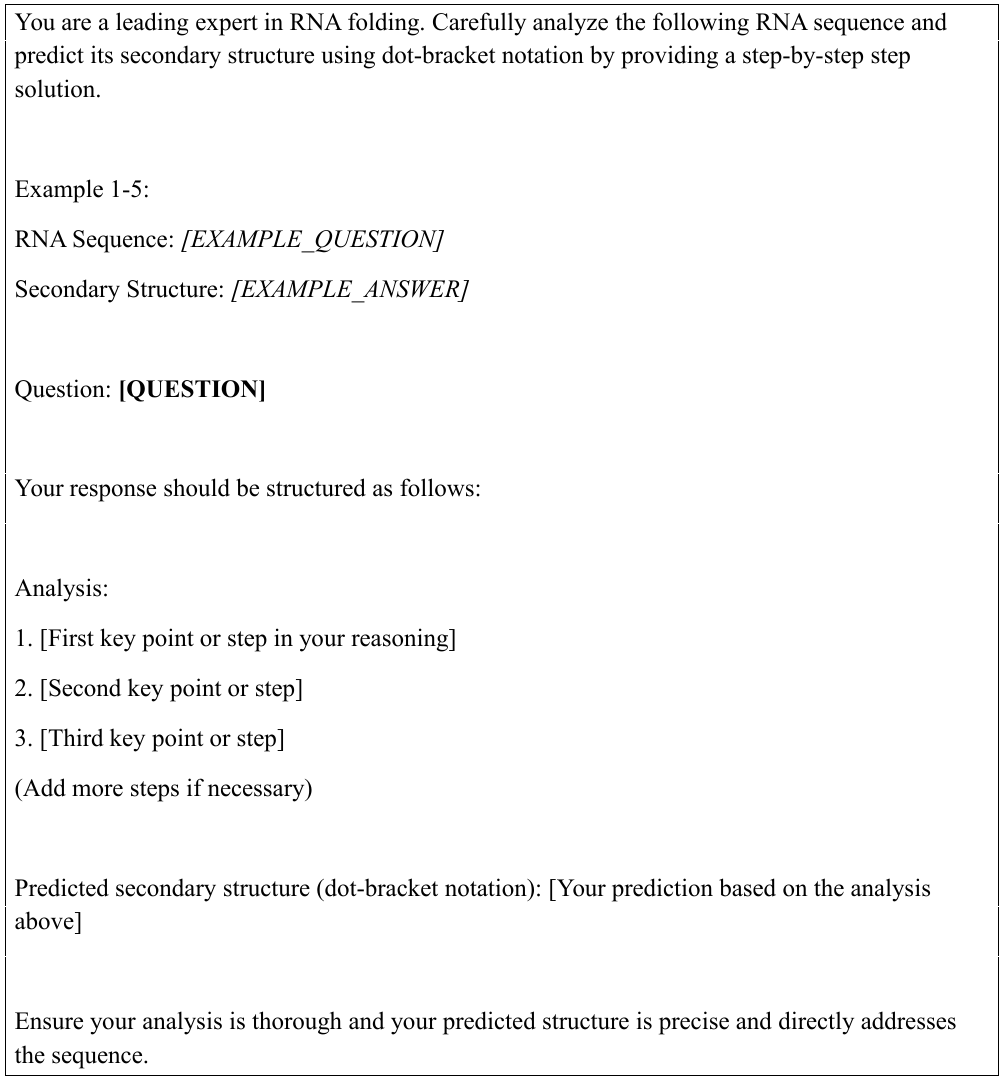} 
\end{figure}
\begin{figure}[h!]
    \centering
    \includegraphics[width=1\textwidth]{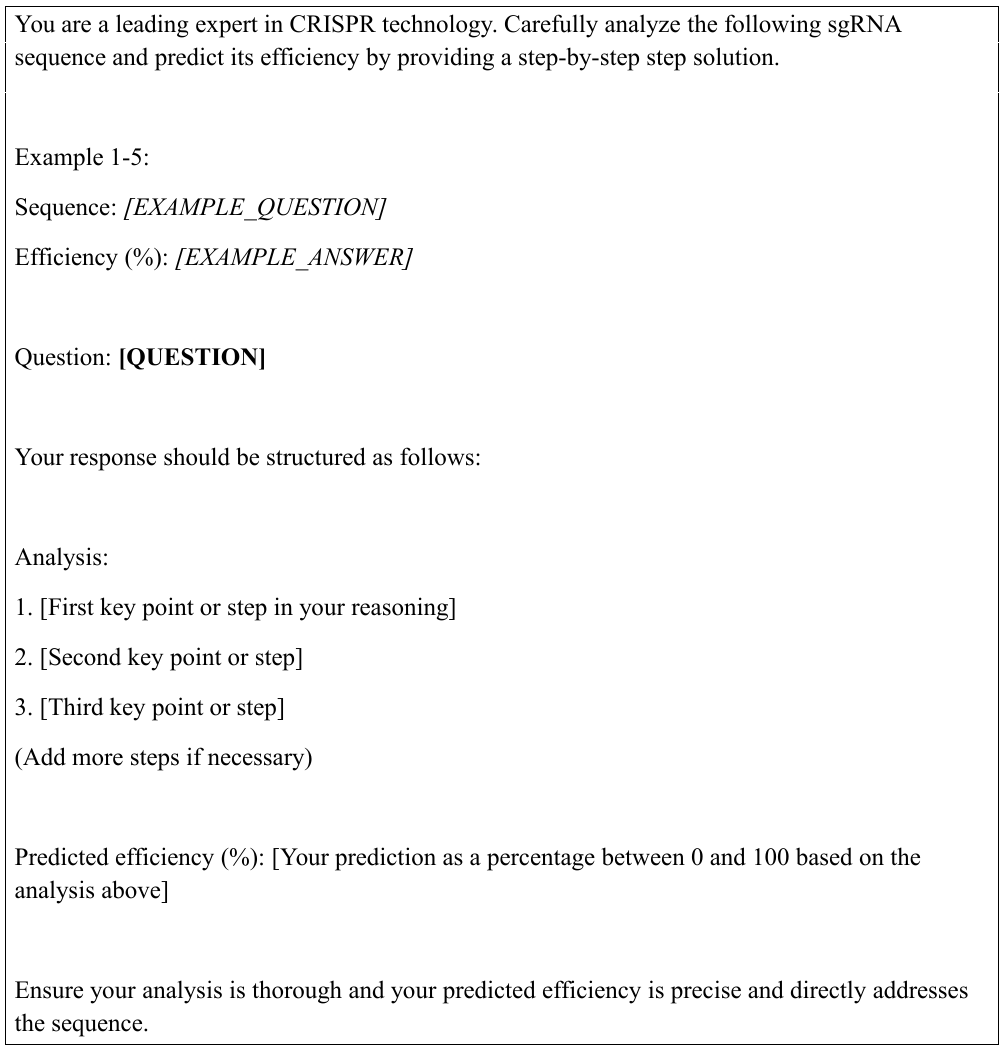} 
\end{figure}
\begin{figure}[h!]
    \centering
    \includegraphics[width=1\textwidth]{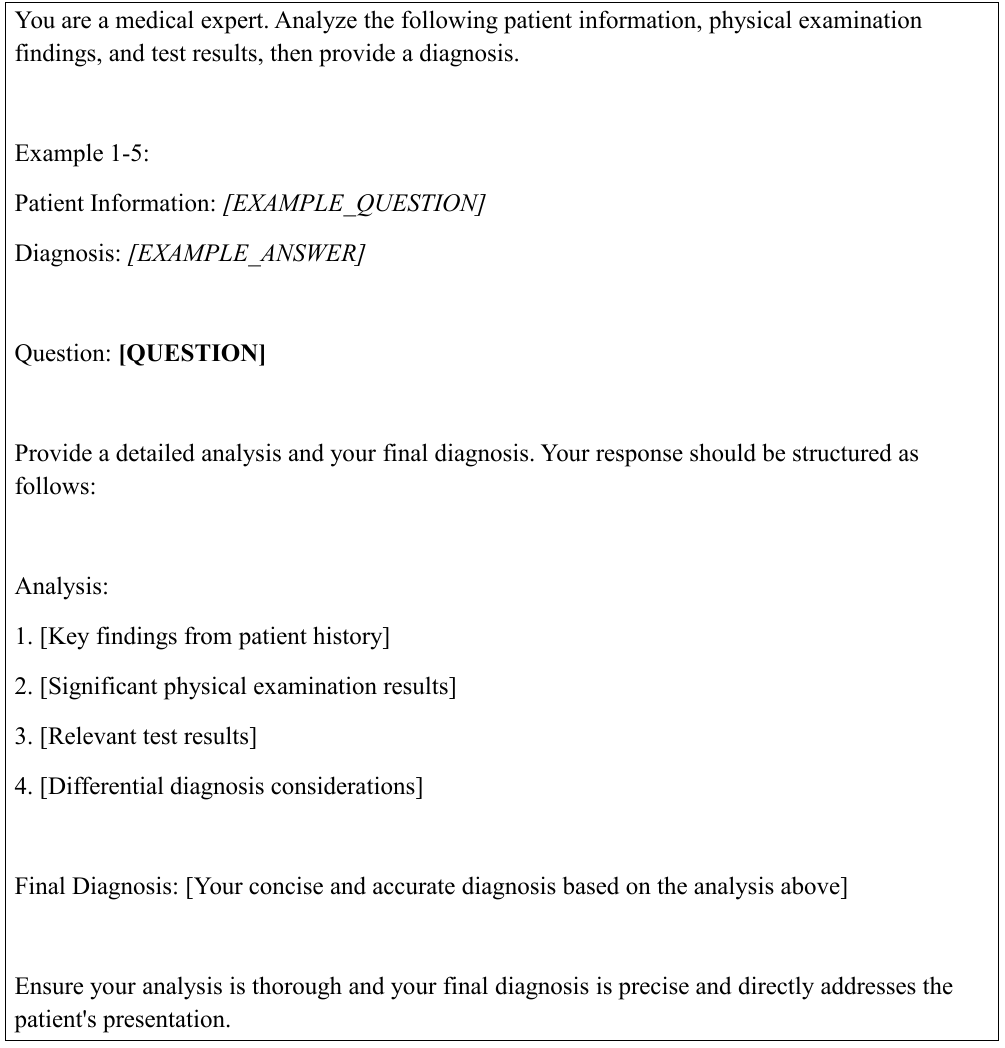} 
\end{figure}
\begin{figure}[h!]
    \centering
    \includegraphics[width=1\textwidth]{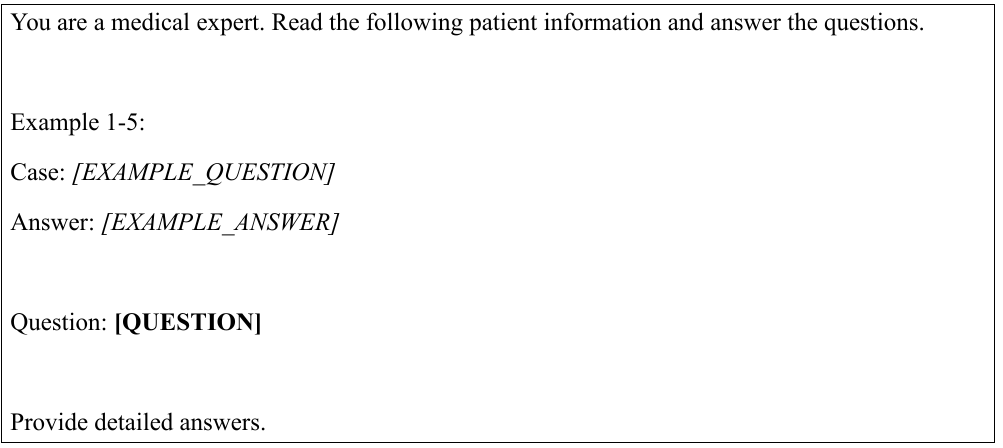} 
\end{figure}
\begin{figure}[h!]
    \centering
    \includegraphics[width=1\textwidth]{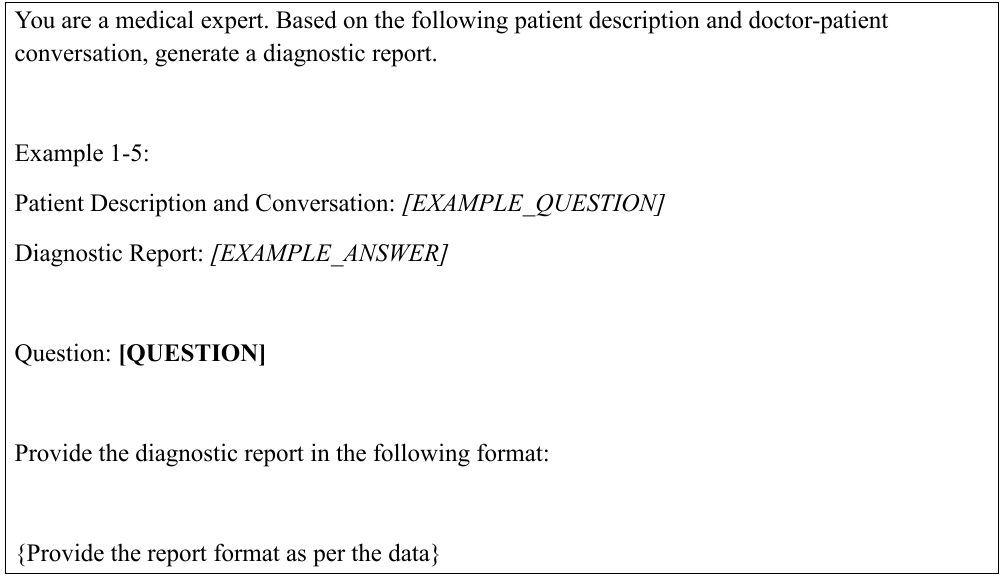} 
\end{figure}
\begin{figure}[h!]
    \centering
    \includegraphics[width=1\textwidth]{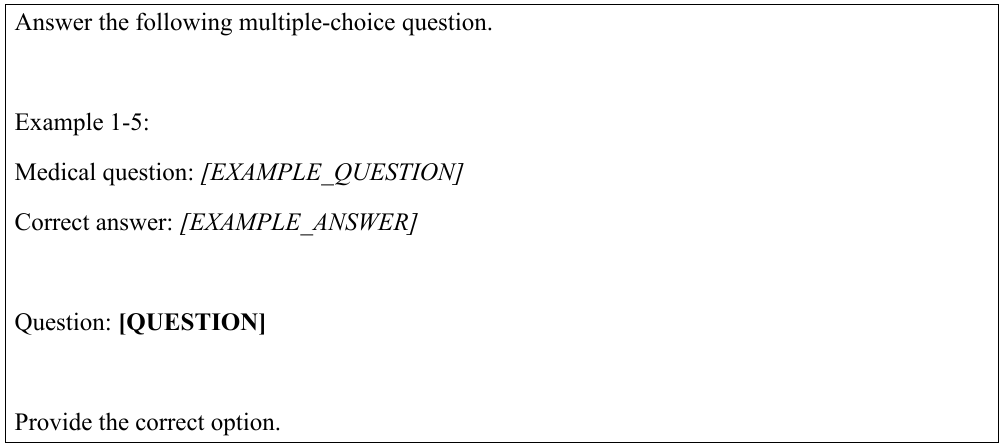} 
\end{figure}
\begin{figure}[h!]
    \centering
    \includegraphics[width=1\textwidth]{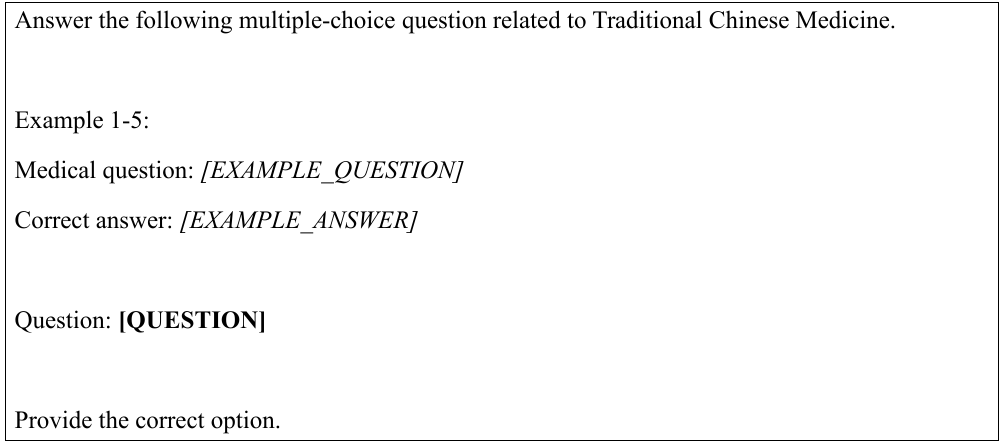} 
\end{figure}
\begin{figure}[h!]
    \centering
    \includegraphics[width=1\textwidth]{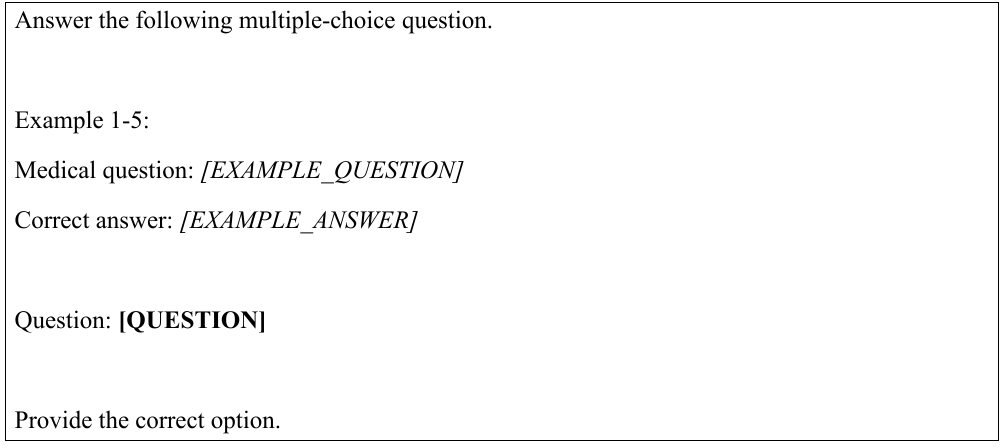} 
\end{figure}
\begin{figure}[h!]
    \centering
    \includegraphics[width=1\textwidth]{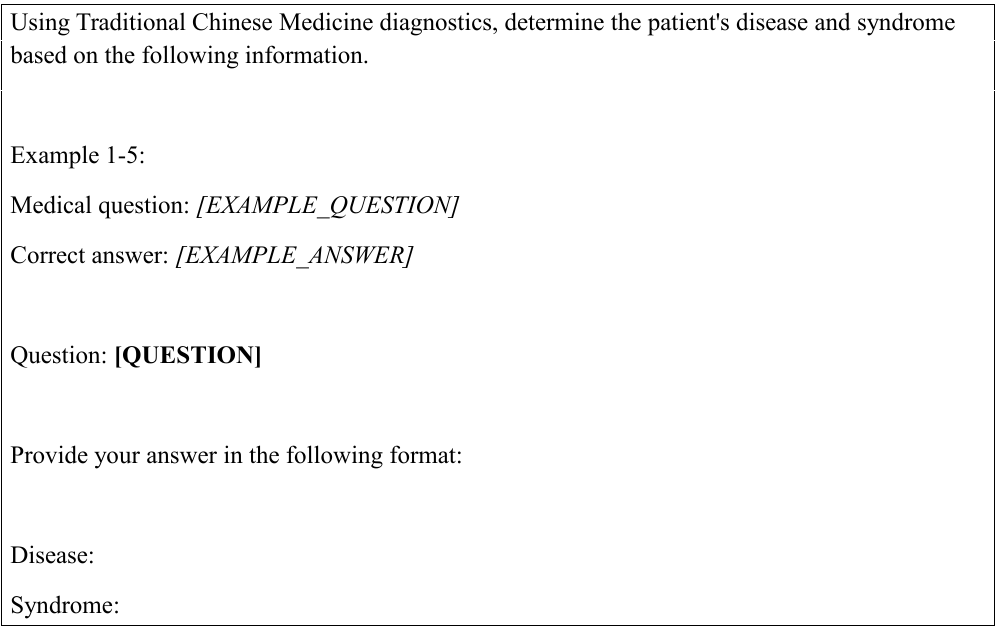} 
\end{figure}
\begin{figure}[h!]
    \centering
    \includegraphics[width=1\textwidth]{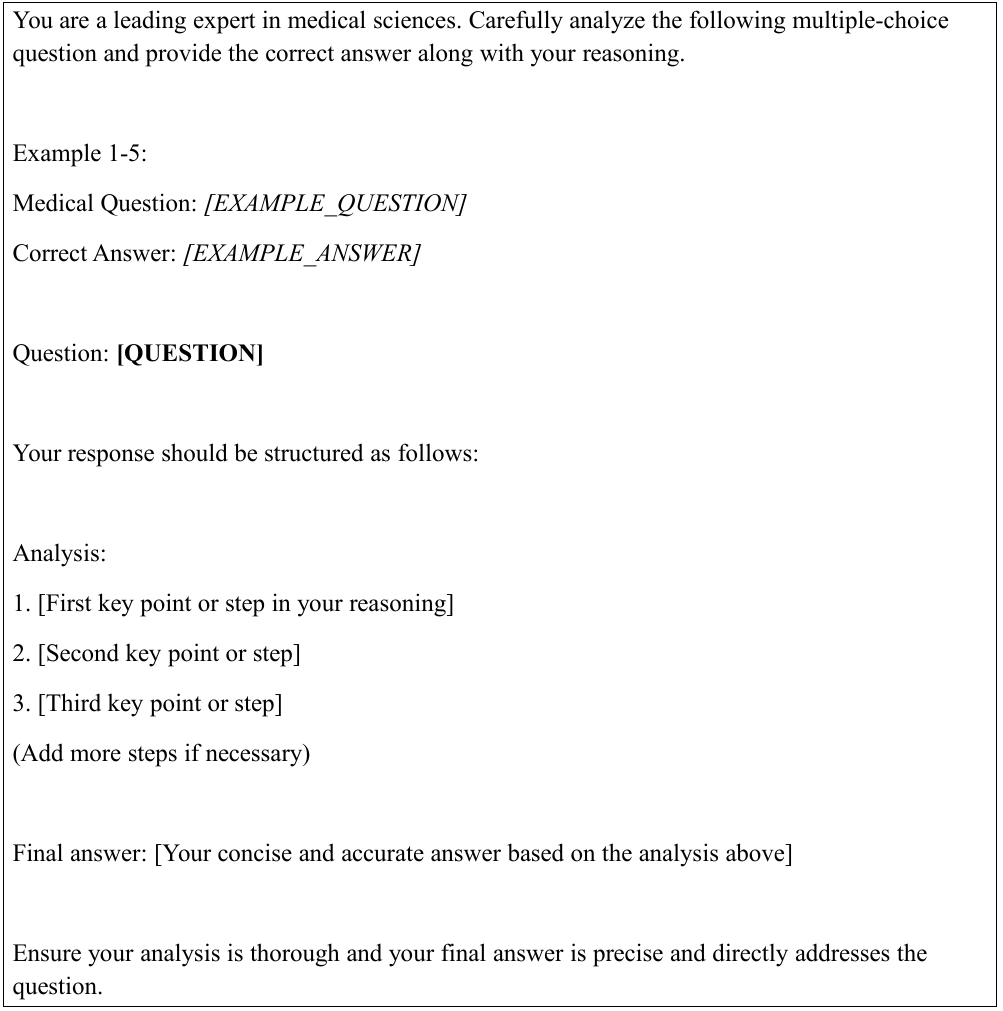} 
\end{figure}
\begin{figure}[h!]
    \centering
    \includegraphics[width=1\textwidth]{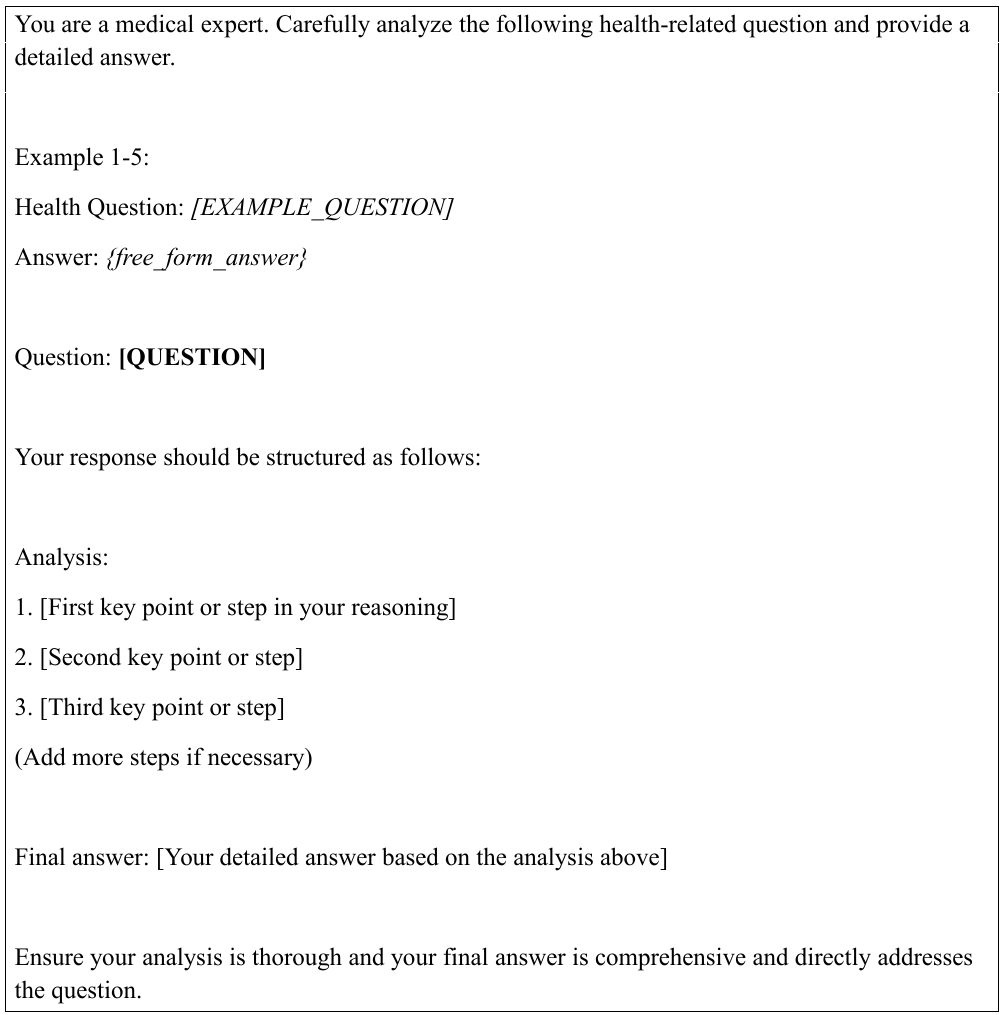} 
\end{figure}
\begin{figure}[h!]
    \centering
    \includegraphics[width=1\textwidth]{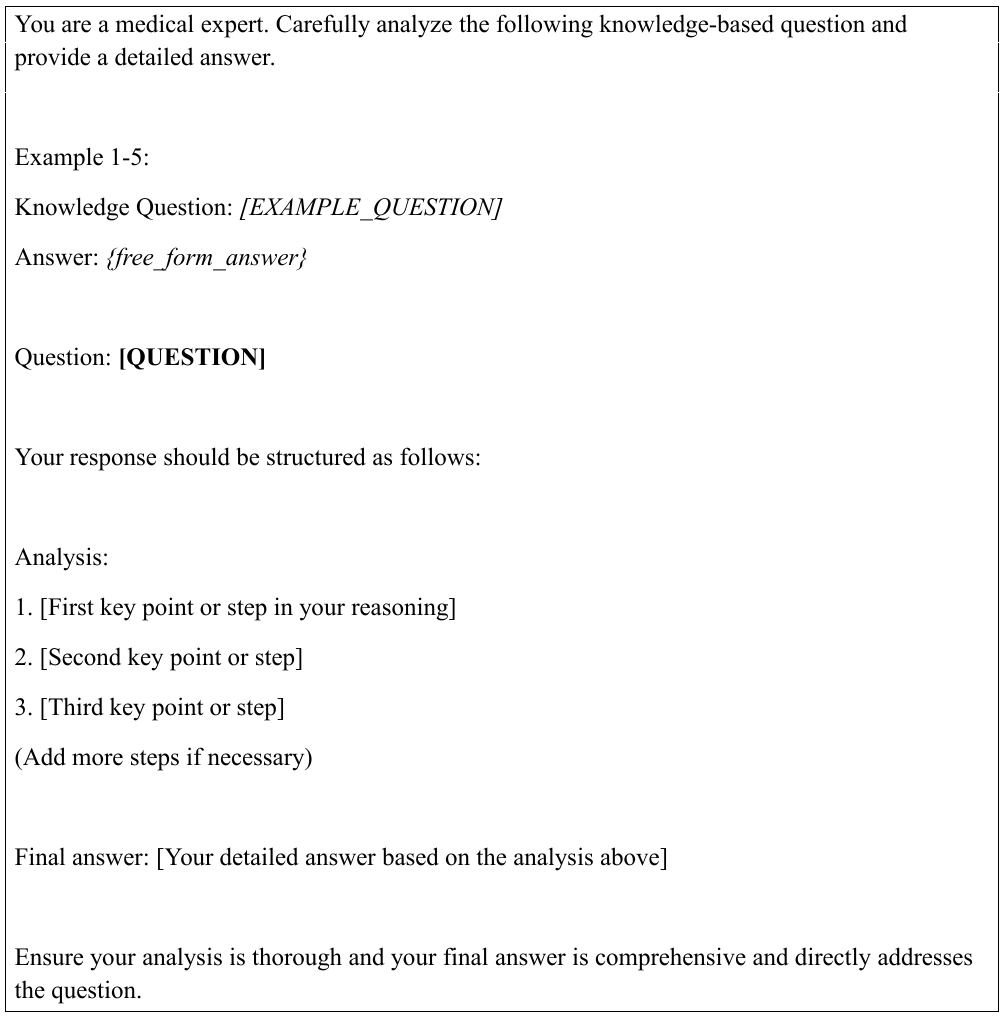} 
\end{figure}
\begin{figure}[h!]
    \centering
    \includegraphics[width=1\textwidth]{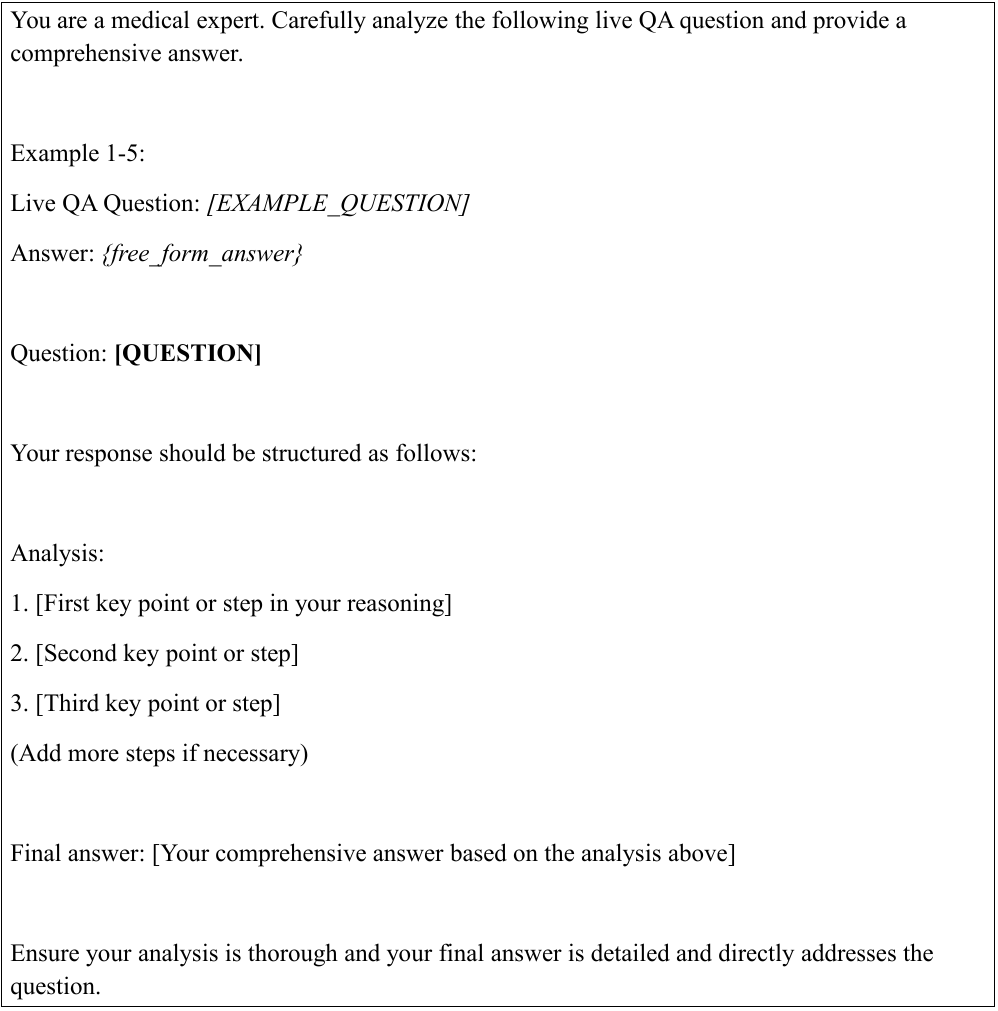} 
\end{figure}
\begin{figure}[h!]
    \centering
    \includegraphics[width=1\textwidth]{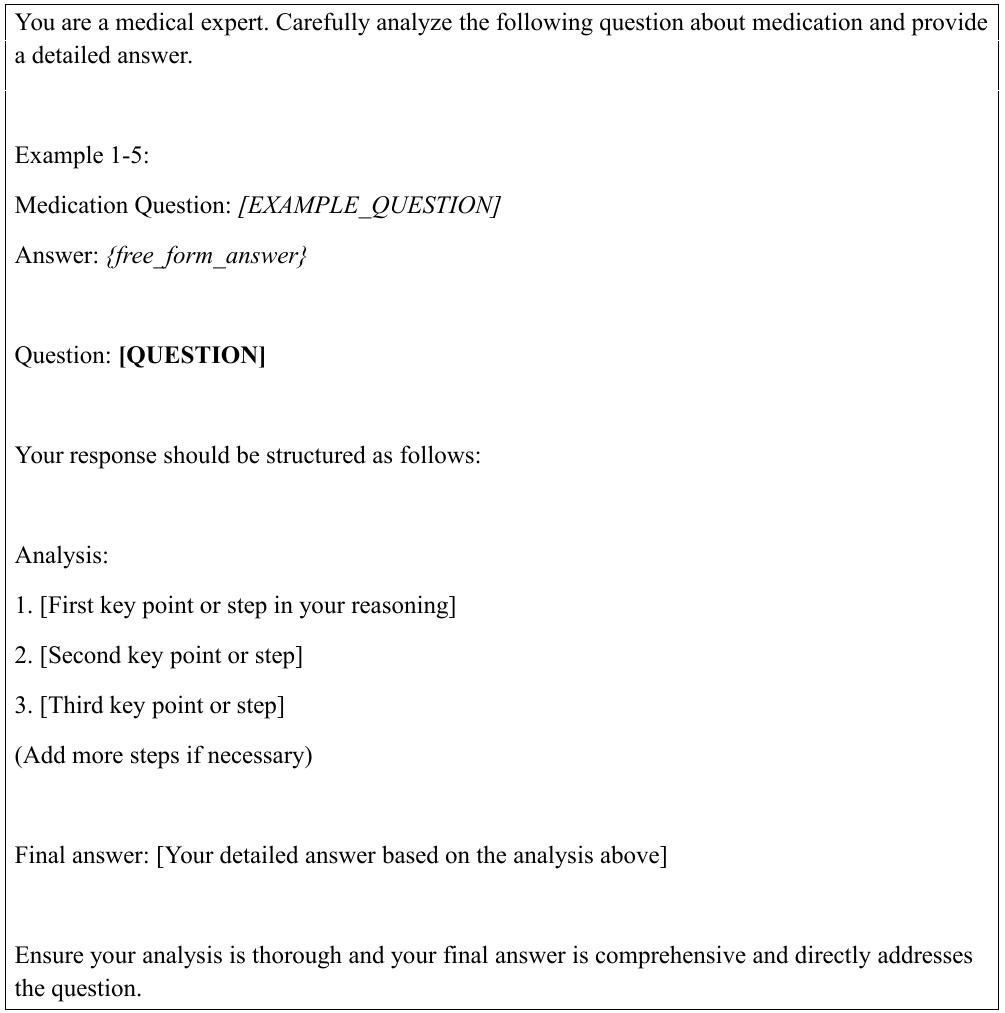} 
\end{figure}
\begin{figure}[h!]
    \centering
    \includegraphics[width=1\textwidth]{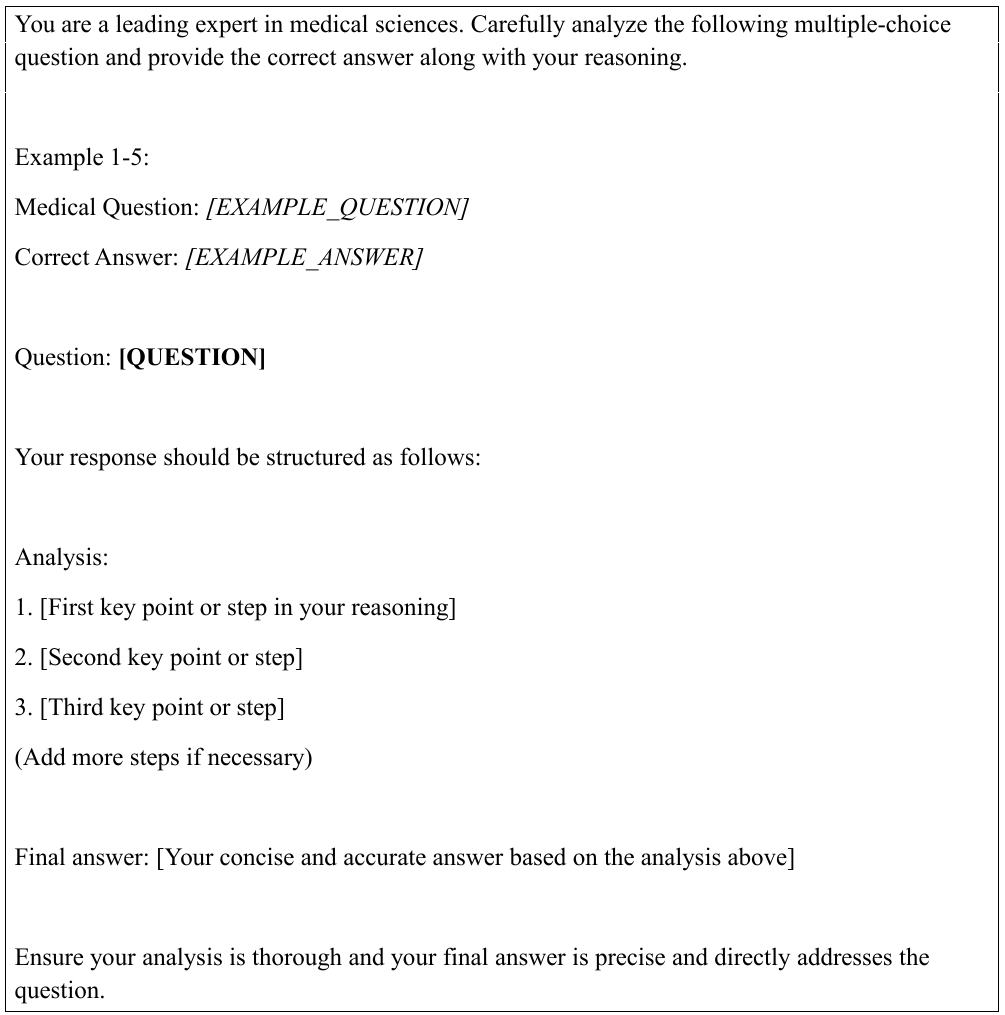} 
\end{figure}
\begin{figure}[h!]
    \centering
    \includegraphics[width=1\textwidth]{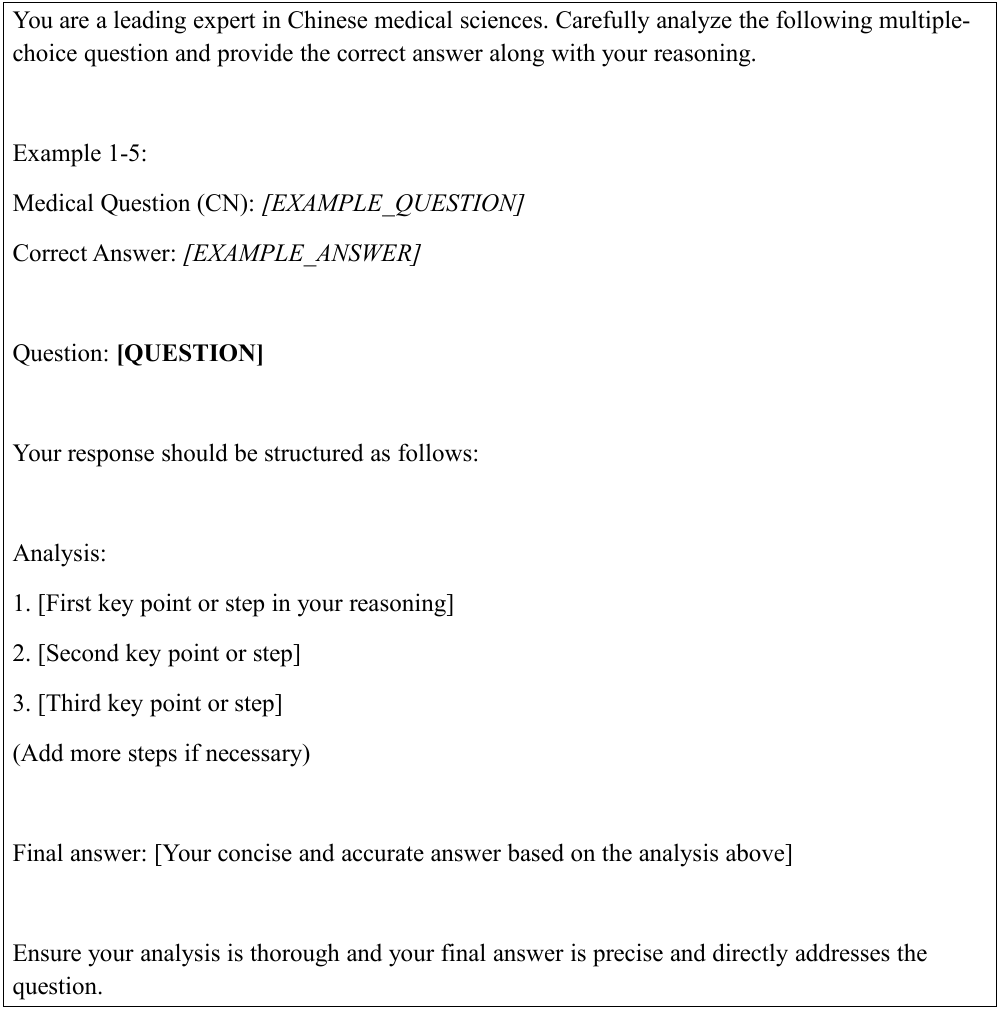} 
\end{figure}
\begin{figure}[h!]
    \centering
    \includegraphics[width=1\textwidth]{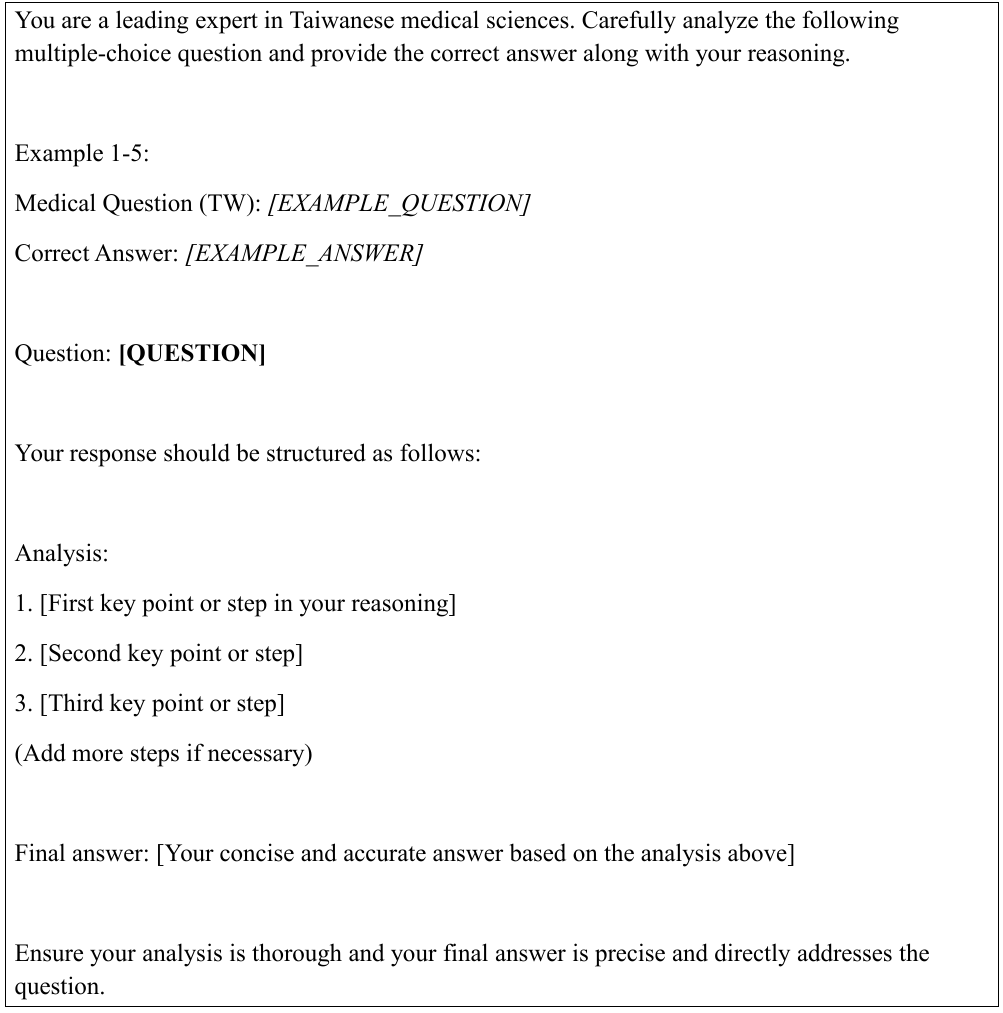} 
\end{figure}
\begin{figure}[h!]
    \centering
    \includegraphics[width=1\textwidth]{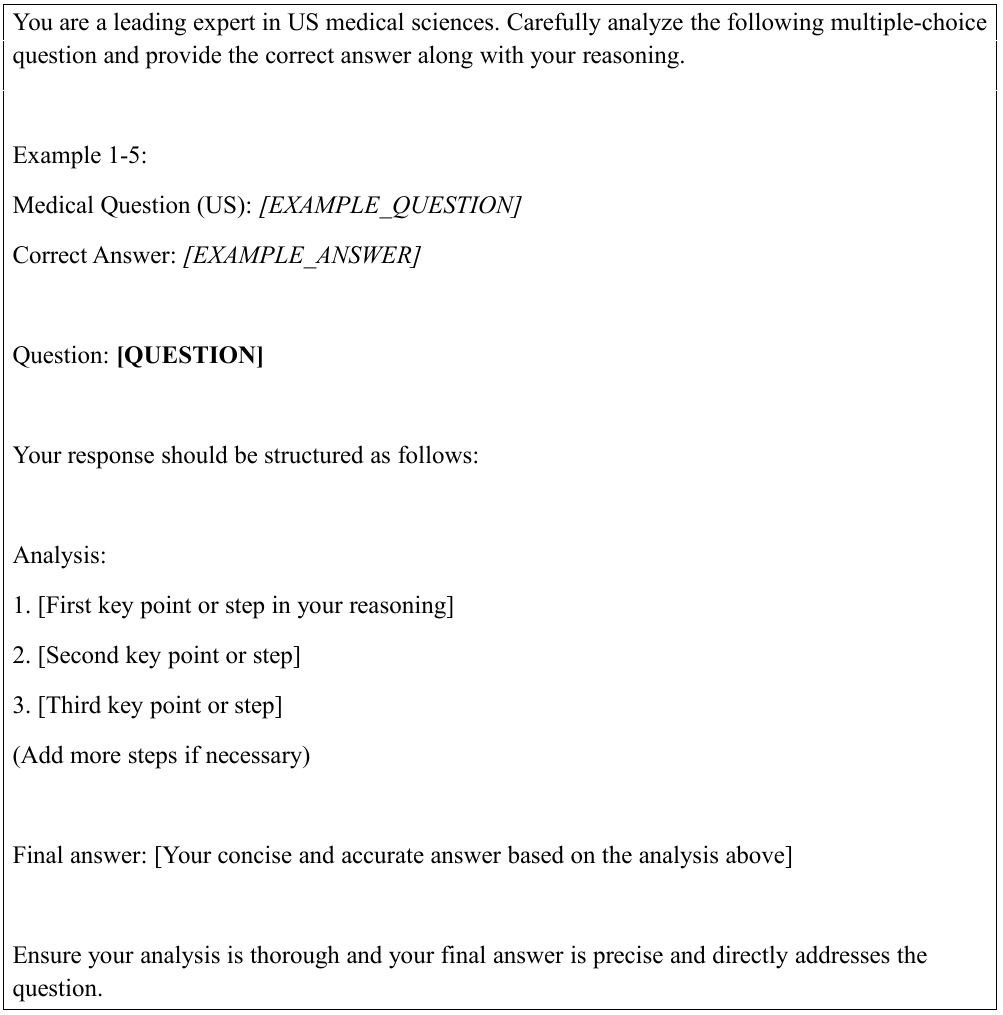} 
\end{figure}
\begin{figure}[h!]
    \centering
    \includegraphics[width=1\textwidth]{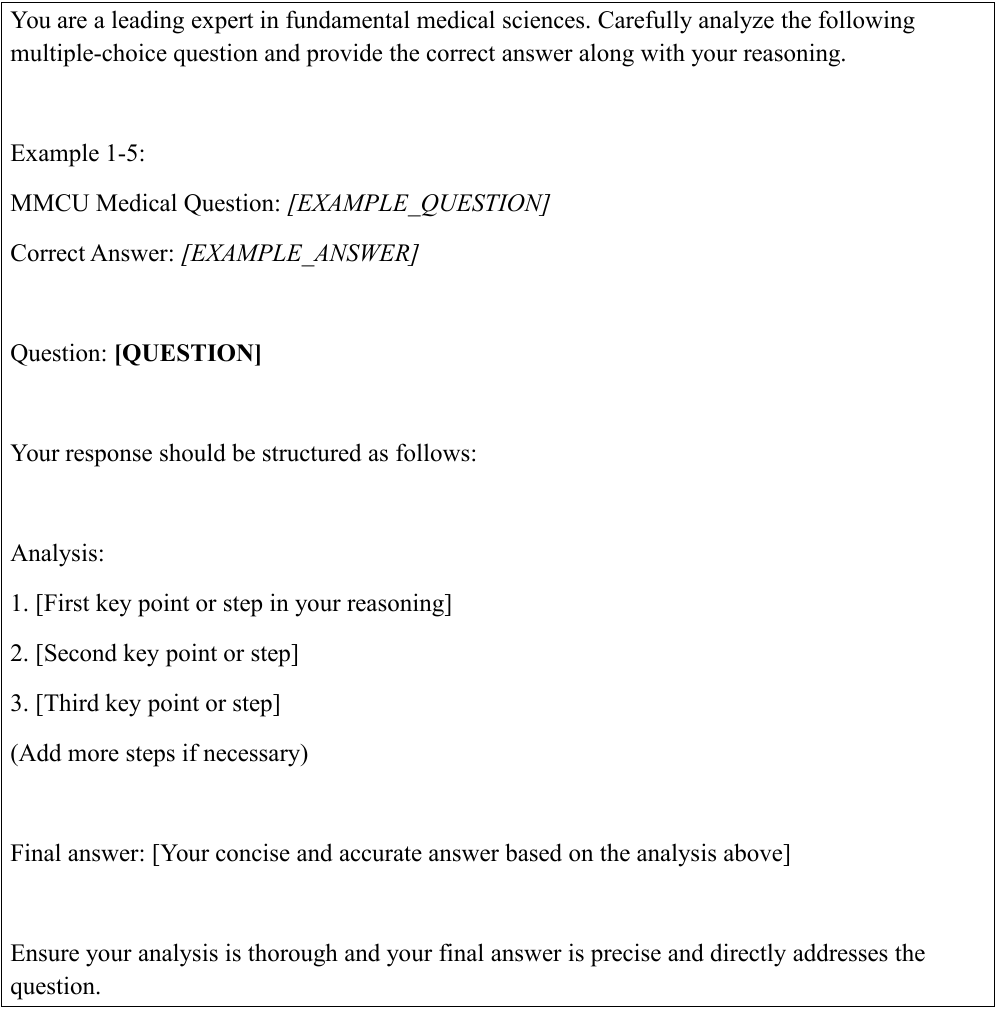} 
\end{figure}
\begin{figure}[h!]
    \centering
    \includegraphics[width=1\textwidth]{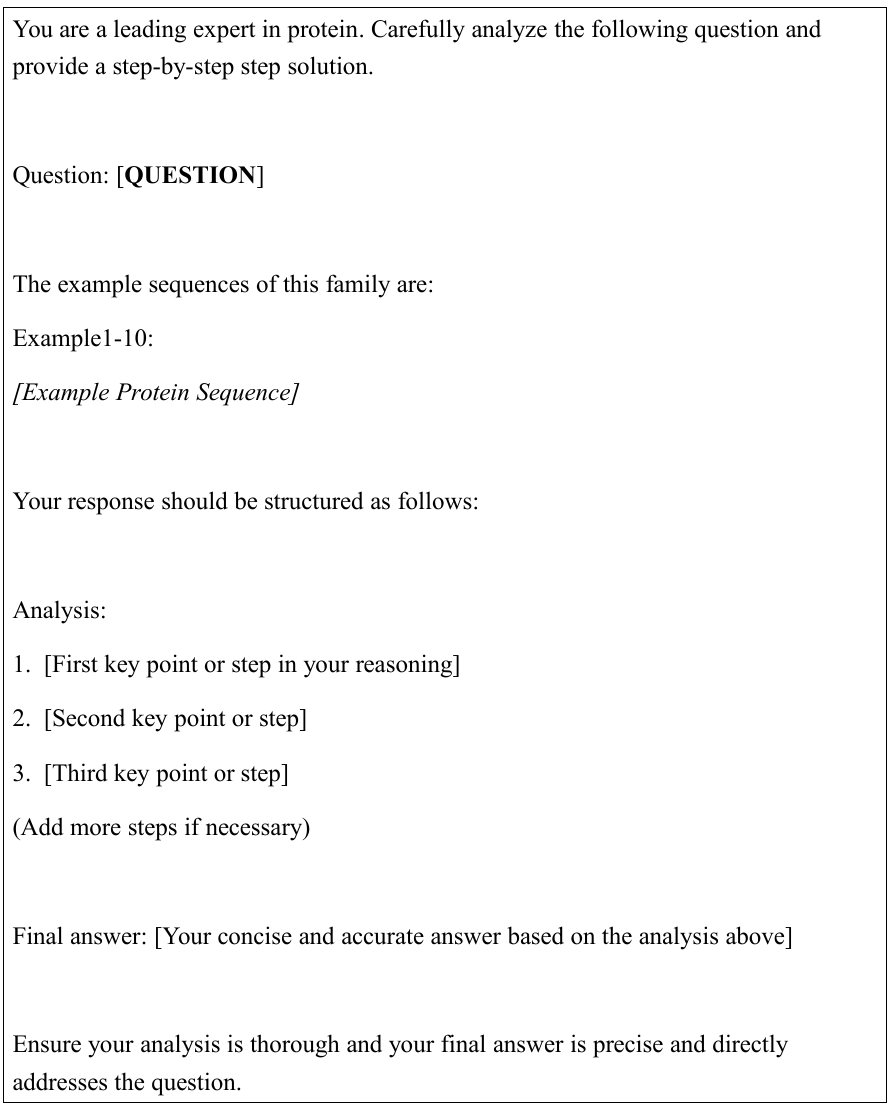} 
\end{figure}
\begin{figure}[h!]
    \centering
    \includegraphics[width=1\textwidth]{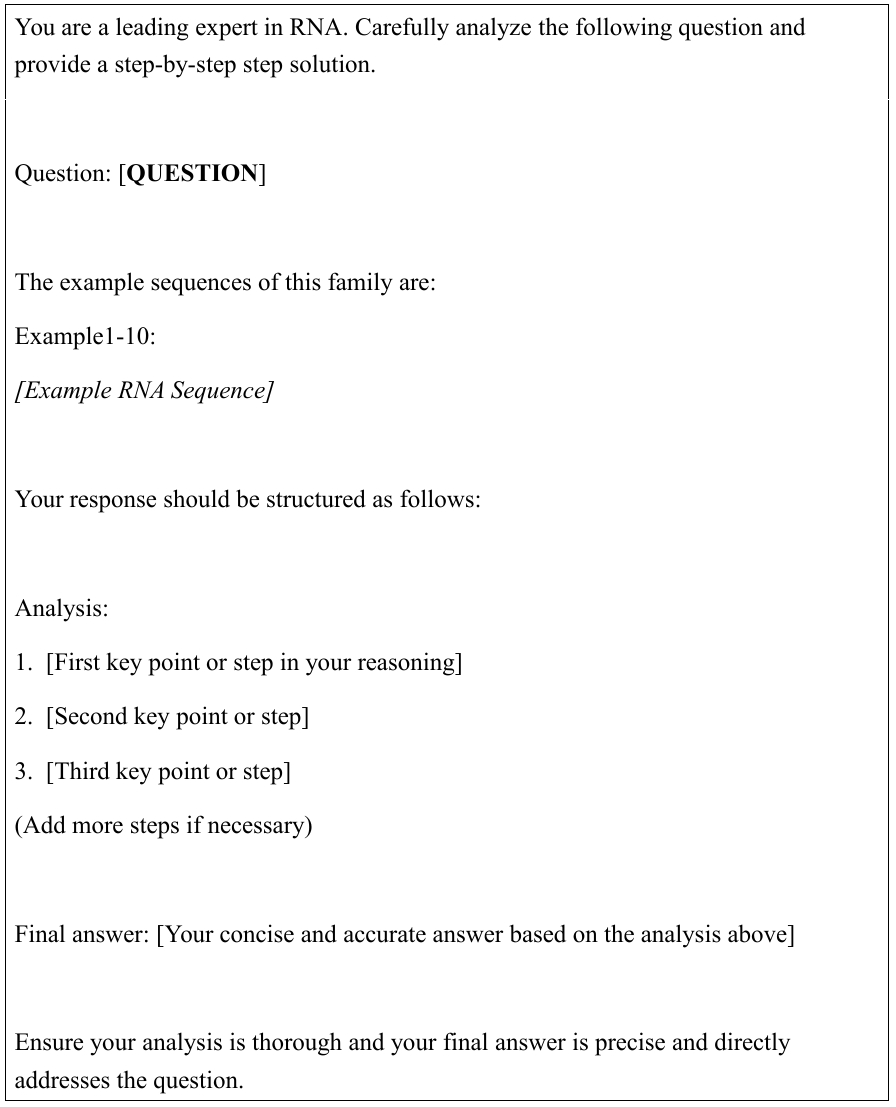} 
\end{figure}

\end{document}